\documentclass[10pt,twocolumn,letterpaper]{article}

\usepackage[pagenumbers]{config/cvpr} %

\usepackage{tikz}
\usepackage{enumitem}
\usepackage{multirow}
\usepackage{adjustbox}
\usepackage{amsmath}

\usepackage{makecell}
\usepackage{pifont}%

\definecolor{cvprblue}{rgb}{0.21,0.49,0.74}
\usepackage[pagebackref,breaklinks,colorlinks,citecolor=cvprblue]{hyperref}

\usepackage{balance}

\usepackage{fontawesome}

\usepackage{lipsum}
\usepackage{tcolorbox}

\usepackage[normalem]{ulem}

\usepackage{xspace}
\usepackage{currfile}
\xspaceaddexceptions{]\}}

\newcommand{\refine}[1]{#1}

\newcommand{\camready}[1]{{#1}\xspace}

\newcommand{\nameCOLOR}[1]{#1} %
\newcommand{\DECO}{\nameCOLOR{\mbox{DECO}}\xspace}
\newcommand{\DAMON}{\nameCOLOR{\mbox{DAMON}}\xspace}
\newcommand{\PHOSA}{\nameCOLOR{\mbox{PHOSA}}\xspace}
\newcommand{\SMPL}{\nameCOLOR{\mbox{SMPL}}\xspace}

\newcommand{\CONTHO}{\nameCOLOR{\mbox{CONTHO}}\xspace}
\newcommand{\Procigen}{\nameCOLOR{\mbox{ProciGen}}\xspace}
\newcommand{\InterCap}{\nameCOLOR{\mbox{InterCap}}\xspace}
\newcommand{\BEHAVE}{\nameCOLOR{\mbox{BEHAVE}}\xspace}

\newcommand{\OSX}{\nameCOLOR{\mbox{OSX}}\xspace}
\newcommand{\ContactEdit}{\nameCOLOR{\mbox{ContactEdit}}\xspace}

\newcommand{\nameCollection}{\mbox{PICO}\xspace}
\newcommand{\nameDataset}{\mbox{\nameCollection-db}\xspace}
\newcommand{\nameMethod}{\mbox{\nameCollection-fit}\xspace}
\newcommand{\nameMethodStar}{\mbox{$\text{\nameMethod}^*$}\xspace}
\newcommand{\nameCollectionLONG}{People In Contact with Objects}

\newcommand{\hoi}{\mbox{{HOI}}\xspace}
\newcommand{\HOI}{\hoi}

\newcommand{\OpenShape}{\mbox{\nameCOLOR{OpenShape}}\xspace}
\newcommand{\PointBind}{\mbox{\nameCOLOR{PointBind}}\xspace}
\newcommand{\Objaverse}{\mbox{\nameCOLOR{Objaverse}}\xspace}
\newcommand{\ObjaverseLVIS}{\mbox{\nameCOLOR{Objaverse-LVIS}}\xspace}
\newcommand{\SAM}{\mbox{\nameCOLOR{SAM}}\xspace}

\newcommand{\human}{h}
\newcommand{\object}{o}
\newcommand{\mask}{m}
\newcommand{\penetration}{p}
\newcommand{\collision}{\penetration}
\newcommand{\contact}{c}
\newcommand{\scale}{s}

\newcommand{\IoU}{\mbox{\nameCOLOR{IoU}}}

\newcommand{\obman}{\mbox{\nameCOLOR{ObMan}}\xspace}
\newcommand{\LEMON}{\mbox{\nameCOLOR{LEMON}}\xspace}

\definecolor{myBlue}{rgb}{0.18, 0.38, 0.82}  %

\newcommand{\outTitleFULL}{PICO: Reconstructing 3D People In Contact with Objects}
\newcommand{\outTitleFULLThin}{\vspace{-0.26em}\outTitleFULL\vspace{-0.26em}}

\newcommand{\video}{\textcolor{magenta}{{video on our website}}\xspace}
\newcommand{\Video}{\textcolor{magenta}{{Video on our website}}\xspace}
\newcommand{\supmat}{{Sup.~Mat.}\xspace}

\newcommand{\highlight}[1]{#1} %

\usepackage{lipsum}

\newcommand{\smpl}{\mbox{\nameCOLOR{SMPL}}\xspace}
\newcommand{\smplx}{\mbox{\nameCOLOR{SMPL-X}}\xspace}
\newcommand{\smplX}{\smplx}

\newcommand{\GPT}{\mbox{\nameCOLOR{GPT-4V}}\xspace}
\newcommand{\HDM}{\mbox{\nameCOLOR{HDM}}\xspace}

\newcommand{\LLM}{\mbox{\nameCOLOR{LLM}}\xspace}

\renewcommand{\etal}{\mbox{et al.}\xspace}
\renewcommand{\ie}{\mbox{i.e.}\xspace}
\renewcommand{\eg}{\mbox{e.g.}\xspace}
\renewcommand{\wrt}{\mbox{w.r.t.}\xspace}

\newcommand{\zheading}[1]{\textbf{#1.}}
\newcommand{\qheading}[1]{\noindent\textbf{#1:}}

\newcommand{\sota}{{state-of-the-art}\xspace}
\newcommand{\SOTA}{\mbox{SotA}\xspace}

\newcommand{\inthewild}{{in-the-wild}\xspace}

\newcommand{\colorRef}[1]{#1} %
\usepackage[capitalize]{cleveref}
\crefname{figure}{\colorRef{Fig.}}{\colorRef{Figs.}}
\Crefname{figure}{\colorRef{Figure}}{\colorRef{Figures}}
\crefname{section}{\colorRef{Sec.}}{\colorRef{Secs.}}
\Crefname{section}{\colorRef{Section}}{\colorRef{Sections}}
\Crefname{table}{\colorRef{Table}}{\colorRef{Tables}}
\crefname{table}{\colorRef{Tab.}}{\colorRef{Tabs.}}
\Crefname{equation}{\colorRef{Equation}}{\colorRef{Equations}}
\crefname{equation}{\colorRef{Eq.}}{\colorRef{Eqs.}}

\newcommand{\objectmesh}{\mathcal{O}}
\newcommand{\humanmesh}{\mathcal{H}}

\newcommand{\contactset}{\mathbb{S}}
\newcommand{\loss}{\mathcal{L}}
\newcommand{\maskVec}{M}

\newcommand{\predobjmask}{\bar{\maskVec}_\object}
\newcommand{\gtobjmask}{{\maskVec}_\object}
\newcommand{\predhumanmask}{\bar{\maskVec}_\human}
\newcommand{\gthumanmask}{{\maskVec}_\human}

\definecolor{lightgray}{gray}{0.97}
\definecolor{lightblue}{rgb}{0.93,0.95,1.0}

\definecolor{GreenColor}{rgb}{0.137,0.573,0.565}
\definecolor{OrangeColor}{rgb}{0.914,0.541,0.0.141}
\definecolor{PurpleColor}{rgb}{0.5,0,0.7}

\newlist{todolist}{itemize}{2}
\setlist[todolist]{label=$\square$}

\newcommand{\patch}{{patch}\xspace}

\newcommand{\cmark}{\color{ForestGreen}\ding{51}}

\newcommand{\xmark}{\color{red}\ding{55}}

\def\paperID{5546}
\def\confName{CVPR}
\def\confYear{2025}

\title{\outTitleFULLThin}

\author{
    Alp\'{a}r Cseke$^{1,2\ast  \ddagger}$    \quad 
    Shashank Tripathi$^{1\ast \dagger}$ \quad
    Sai Kumar Dwivedi$^{1}$ \quad
    Arjun S. Lakshmipathy$^{3}$ \quad \\
    Agniv Chatterjee$^{4}$\quad
    Michael J. Black$^{1}$ \quad
    Dimitrios Tzionas$^{5}$ \quad 
    \vspace{0.5em}\\
    \normalsize $^1$Max Planck Institute for Intelligent Systems, T\"{u}bingen, Germany \quad
    \normalsize $^2$Meshcapade\\
    \normalsize $^3$Carnegie Mellon University, USA \quad
    \normalsize $^4$UT Austin, USA \quad
    \normalsize $^5$University of Amsterdam, the Netherlands \\
}

\begin{document}

\twocolumn[{%
\renewcommand\twocolumn[1][]{#1}%
\maketitle
\begin{center}
    \centering
    \vspace{-0.5 em}
    \captionsetup{type=figure}
    \includegraphics[width=0.90 \linewidth]{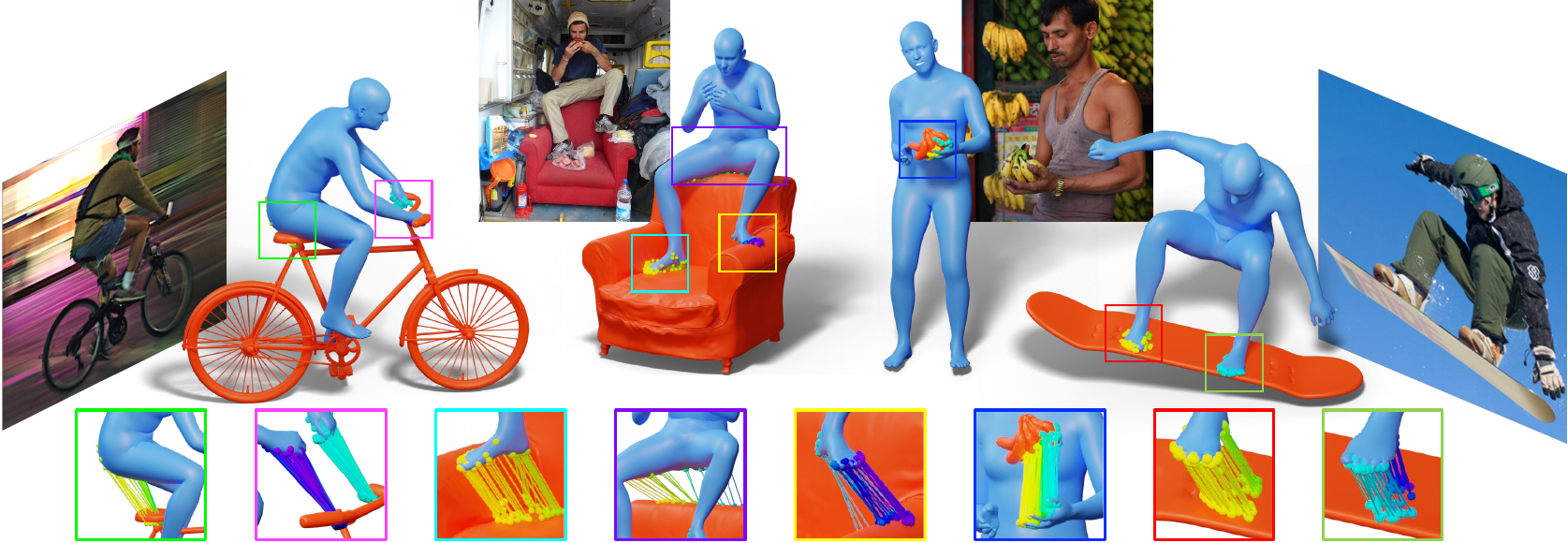}
    \vspace{-0.5 em}
    \caption{
        We present \nameCollection, a novel framework for joint human-object reconstruction in 3D. \nameCollection includes \nameDataset, a unique dataset that pairs natural images with dense vertex-level 3D contact 
        \camready{correspondences on both the human and the object.}
        We leverage this dataset for building \nameMethod, an 
        \highlight{optimization-based %
        method} 
        that fits 3D body and object meshes to an image \highlight{guided by rich contact constraints}. 
        Here, we show reconstruction results of \nameMethod: 
        3D human pose and shape (shown with \textcolor{cvprblue}{blue color}), 3D object pose and shape (shown with \textcolor{BurntOrange}{orange color}), and 
        contact correspondences (shown with 
        \textcolor{ForestGreen}{va}\textcolor{RedViolet}{ri}\textcolor{Violet}{ous} \textcolor{GreenYellow}{co}\textcolor{RoyalPurple}{lo}\textcolor{cyan}{rs} in inset). 
        Note that \nameMethod works for in-the-wild images, 
        as well as for many previously untackled object classes. 
    }
    \label{figure:pico:teaser}
\end{center}%

}]

\def\thefootnote{*}\footnotetext{Equal contribution.
\hspace{0.5cm}\textsuperscript{$\dagger$}Project lead.
\hspace{0.5cm}\textsuperscript{$\ddagger$}Work done while at MPI.}

\begin{abstract}
Recovering 3D Human-Object Interaction (\HOI) from single images is challenging due to depth ambiguities, occlusions, and 
the huge variation in 
object shape and appearance. %
Thus, past work 
requires controlled settings such as known object shapes and contacts, and tackles only limited object classes. 
Instead, we need methods that generalize to natural images and novel object classes. 
\mbox{We tackle this in two main ways}: 
\highlight{(1)}~We collect \nameDataset, a new dataset of natural images uniquely paired with dense 3D contact 
\camready{correspondences} 
on both body and object meshes. 
To this end, we use images from the recent \DAMON dataset that are paired with 
\camready{annotated} 
contacts, but only on a \highlight{canonical} 3D body.
In contrast, we seek contact labels on both the body and the object.
To infer these, given an image, we retrieve an appropriate 3D object mesh from a database by leveraging vision foundation models. 
Then, we project \DAMON's body contact patches onto the object via a novel method needing only 2 clicks per patch. 
This minimal human input establishes rich contact correspondences between bodies and objects. 
\highlight{(2)}~We exploit 
our new \camready{dataset} 
in a novel \highlight{render-and-compare fitting method}, called \nameMethod, to recover 3D body and object meshes in \highlight{interaction}. %
\nameMethod infers contact for the \smplX body, %
retrieves a likely 3D object mesh \highlight{and contact} from \nameDataset for that object, and 
uses the contact to \highlight{iteratively fit} 
the 3D body and object meshes to image evidence \highlight{via optimization}. 
Uniquely, \nameMethod 
works well for 
many object classes that no existing method can tackle. 
This is crucial \camready{for scaling} \HOI understanding \camready{in the wild}. 
Our 
data and code 
\camready{are available} at 
\url{https://pico.is.tue.mpg.de}. 
\vspace{-1.0 em}
\end{abstract}

\section{Introduction}
\label{sec:intro}

Humans routinely interact with 
objects. 
Thus, recovering Human-Object Interaction (\HOI) in 3D from natural images is important for human-centric applications such as smart homes, 
mixed reality, 
or assistive robots. 
At its core, this entails inferring human pose and shape, object pose and shape, and their spatial arrangement and contacts, all in 3D. 

Despite progress, %
the field lies at its infancy due to strong challenges; humans and objects come in a huge variety of shapes, they mutually occlude each other, 
and contact is often ambiguous in 2D images.
Thus, most work focuses on controlled settings, with known object shapes or contacts. 
To be practical, however, we need to infer 3D \HOI %
from \camready{unconstrained} 2D images \camready{taken} in the wild.

For this task, 
current methods struggle for two reasons. 
\highlight{First}, %
no method robustly recovers 
3D object shape from single images 
because,
unlike \camready{for} human bodies, 
there 
\camready{exists no}
single statistical model for object shape. 
And while we might hope that foundation models would provide a solution, their 3D reasoning skills are still limited.
\mbox{\highlight{Second}}, 
given 3D body and object shapes in an image, 
no method robustly recovers their 3D pose %
and arrangement. 
Knowing the contact between the body and the object would facilitate pose estimation of both.
Unfortunately, current methods that regress contact information from images 
either  (i) infer contact only in 2D~\cite{chen2023hot}, (ii)
infer 3D contacts only on the body~\cite{tripathi2023deco} ignoring objects, or (iii) 
train on synthetic 
data 
\cite{shimada2022hulc} 
\camready{so they struggle generalizing} 
to real images.

We tackle these key limitations with a novel framework called \nameCollection ({``\nameCollectionLONG''}) which has three key properties:
\highlight{(1)}~It facilitates 3D \HOI reasoning in \emph{natural images} %
with widely varying viewpoints, occlusions, body poses, and objects. %
\highlight{(2)}~It supports human interaction with \emph{arbitrary object classes}, 
without requiring an a-priori known object type or shape. %
\highlight{(3)}~It enables the detection of \emph{dense contacts} %
on \emph{both} the human and the object that
establish rich point \emph{correspondences} between them.

Specifically, our \nameCollection approach introduces two novelties:
\highlight{(1)}~\nameDataset, a dataset of natural images uniquely paired with dense body-object 3D contact annotations, and 
\highlight{(2)}~\nameMethod, a novel method for reconstructing accurate 
3D \HOI from natural color images by exploiting rich contacts. 
We collect \nameDataset and develop \nameMethod as follows.

\begin{figure} %
    \centering
    \vspace{-0.3 em}
    \includegraphics[width=\linewidth]{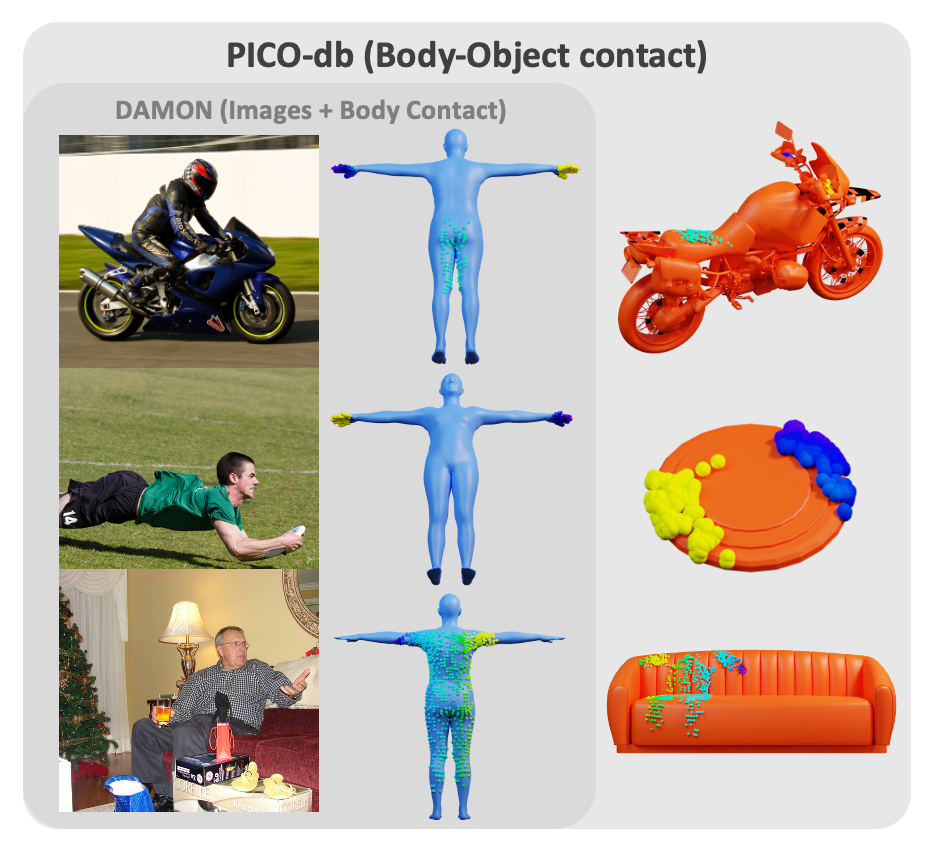}
    \caption{
        \textbf{\nameDataset dataset annotations.} 
        Left to right:
        Color image. 
        Contacts 
        (shown in \textcolor{ForestGreen}{va}\textcolor{RedViolet}{ri}\textcolor{Violet}{ous} \textcolor{GreenYellow}{co}\textcolor{RoyalPurple}{lo}\textcolor{cyan}{rs}) 
        annotated on the \textcolor{cvprblue}{body}
        and \textcolor{BurntOrange}{object}.
        Contact annotations establish bijective body-object correspondences, denoted with color-coding. 
    }
    \label{fig:dataset:pico_dataset_examples}
\vspace{-0.5 em}
\end{figure}

\zheading{\nameDataset:~3D \HOI contacts}
To train models that infer 3D %
contacts from 
\inthewild 
images, we need data. %
The only such dataset is \DAMON~\cite{tripathi2023deco}, which pairs images with 
3D contacts on the body. 
These annotations are
crowd-sourced 
via an online tool where people ``paint'' on a T-posed 3D body the contact points present in an image; see \cref{fig:dataset:pico_dataset_examples} (dark-gray box). 
However, \DAMON ignores objects. 
This is a key limitation. %
Moreover, it is non-trivial to extend this painting tool to include annotating contact on objects.
In particular, one needs to ensure that the contacts ``painted'' on an object  agree with those ``painted'' on the body.

Therefore, to build \nameDataset, we
repurpose 
\DAMON's body contacts for objects, inspired by the \ContactEdit~\cite{lakshmipathy2023contactedit} method. 
To this end, we observe that body contacts form ``patches'' of neighboring vertices. 
\ContactEdit defines a finite-length ``axis'' per contact \patch 
as a fine-grained control for translating, rotating, and deforming it. %
Crucially, it lets us transfer the \patch onto another mesh by just re-drawing the axis onto the latter. 
However, %
this is intuitive only for experts. 
Instead, we need to %
democratize this for non-experts to \camready{collect} %
data at scale.
To this end, we 
automatically generate 
an axis per \patch via PCA; the first principal component of contact point locations provides the axis direction, and along this 
we sample its start-/end-ing points. %
Crucially, this means that projecting a contact \patch onto a new mesh requires just two clicks to define the (auto-created) axis. 
Since this is 
an easy 
task, we integrate it into an online tool and use AMT~\cite{amtWEB} to crowd-source 
3D object annotations for \DAMON's images and 3D body contacts.

\camready{A key} 
problem remains, however. 
This annotation process requires a 3D object mesh that is both detailed and manifold. 
To automate the estimation of 3D object shape, we exploit
a large-scale database (we curate \Objaverse~\cite{deitke2023objaverse}) and the recent \OpenShape~\cite{liu2023openshape} foundation model. 
The latter 
embeds both images and point clouds (or meshes by extension) in a single latent space. 
At test time, we embed an image in the latent space, find the nearest-neighbor latent code, and retrieve the respective mesh that likely matches the image. 
This is a simple-yet-efficient, scalable solution. 

Using this approach, we collect \nameDataset, which contains natural images with %
3D contact annotations for both humans and objects; see \cref{fig:dataset:pico_dataset_examples}. 
Note that our contact transfer is almost-isometry preserving, \ie, \nameDataset has bijective body-object correspondences (color-coded in \cref{fig:dataset:pico_dataset_examples}).

\zheading{\nameMethod:~3D \HOI from a 2D image} %
We develop a new method, called \nameMethod, that 
takes in a natural image %
and recovers 3D 
human pose and shape, %
object pose and shape, %
and their %
spatial arrangement. 
To this end, we employ an optimization-based render-and-compare fitting method. 
Specifically, we first initialize 3D body shape and pose via the \OSX~\cite{lin2023osx} model. 
Then, we initialize 3D object shape via \OpenShape-based 
database 
retrieval (see \nameDataset paragraph above), which scales to novel classes. 
However, initializing object pose in 3D \wrt %
the body is challenging. 
We solve for 3D object pose by exploiting \nameDataset's body-object \emph{contact point correspondences} as follows.

When operating on \nameDataset images we simply exploit its annotations. %
But when operating on unlabeled images, there exist no contact correspondences, so past work handcrafts these~\cite{zhang2020phosa}. %
Instead, we automatically infer these. 
To this end, 
given an image, 
we first infer 3D body contacts using \DECO~\cite{tripathi2023deco}, and
the object class using \SAM~\cite{kirillov2023segmentAnything}. 
Based on these, we then
retrieve from \nameDataset the nearest-neighbor body contacts, 
and the respective object shape, object contacts, and body-object contact correspondences. 
\mbox{We find this simple} %
approach to be surprisingly effective %
and we demonstrate how contact correspondences aid in 3D recovery of humans interacting with objects (see \cref{figure:pico:teaser}).

\zheading{Evaluation}
We extensively compare \nameMethod, both quantitatively and qualitatively, with 
\sota methods 
(\PHOSA \cite{zhang2020phosa}, \HDM~\cite{xie2023tprocigen}, \CONTHO~\cite{nam2024contho}). 
A perceptual study shows that 
\nameMethod reconstructions 
are perceived as much %
more realistic. 
Applying \nameMethod on unlabeled images shows that it performs well for many previously untackled object classes, \eg, couches, bananas, and frisbees, %
demonstrating its ability to scale.

In summary, we make the following main contributions:
\begin{enumerate}
    \item 
    We collect \nameDataset, the first dataset of natural images paired with 3D contact on \emph{both} humans and objects, with dense bijective contact \emph{correspondences} between them. 
    \item 
    To build \nameDataset we develop a new method that projects existing body contacts onto objects with minimal effort. 
    \item 
    We build \nameMethod, a method that recovers 3D \HOI from an image, scaling to previously untackled object classes. 
\end{enumerate}
Data and code 
\camready{are available at} 
\href{https://pico.is.tue.mpg.de}{https://pico.is.tue.mpg.de}.

\section{Related Work}
\label{sec:relatedwork}

\subsection{3D Humans from single images}
\label{sec:related:humans-alone}

Estimating 3D human pose and shape from single images has 
evolved 
from optimization- %
to learning-based methods. %
Optimization-based methods fit a parametric model~\cite{SMPL:2015, SMPL-X:2019, xu2020ghum} to image cues such as keypoints~\cite{smplify, xu2020ghum, SMPL-X:2019}, silhouettes~\cite{nbf, dsr}, or 
body-part segmentation masks~\cite{unitepeople}. 
Learning-based methods directly infer body-model parameters from images~\cite{Kocabas2021pare, li2021hybrik, poco, rong2021frankmocap, li2022cliff, tripathi2023ipman} or videos~\cite{VIBE:CVPR:2020, Kanazawa2019Learning3H}. 
However, some methods infer bodies in model-free fashion as vertices~\cite{meshgraphormer,cmr,metro} or via implicit functions~\cite{pifuhd,Mihajlovic_CVPR_2022,xiu2023econ}. 
Recent work~\cite{goel2023humans, lin2023osx, tokenhmr} uses transformers 
for robust inference; here we use the \OSX~\cite{lin2023osx} model.

\subsection{3D Objects from single images}
\label{sec:related:objects-alone}

The field has extensively studied estimating 3D objects from images. 
\camready{To this end, it has used} 
explicit 3D representations \camready{such as} voxels~\cite{imgto3d_voxel1, imgto3d_voxel2}, point clouds~\cite{imgto3d_pointclouds} or meshes~\cite{imgto3d_mesh}, 
\camready{but recently also} implicit representations %
to represent objects of varied topologies \cite{imgto3d_nerf1, imgto3d_nerf2, antic2025sdfit}. 
\camready{Here,} we focus on recent learning-based methods made possible
by 3D object-shape datasets~\cite{wu2015modelNet, xiang2014pascal3D}, \camready{as} 
a \camready{detailed} review is beyond our scope. 
However, 
\camready{such} 
methods can only tackle limited object categories present in training datasets.

Recent work %
goes beyond 
limited categories by using %
text-to-image diffusion models~\cite{imagen, stablediffusion} and large-scale 3D datasets~\cite{deitke2023objaverse, deitke2023objaverseXL}. 
\mbox{Zero-1-to-3}~\cite{zero1to3} %
re-trains a 2D diffusion model to build a viewpoint-conditioned 3D diffusion model. 
Others 
combine 
2D and 3D diffusion models~\cite{qian2023magic123}. 
Despite promising results, 
all these methods require objects to be unoccluded in images, which is 
unrealistic 
for \hoi. 
While text-to-3D models~\cite{magic3d, dreamfusion} %
do not 
have this problem, %
accurately describing objects in text is often difficult. %

To address these issues, we harness recent foundation models \cite{girdhar2023imageBind} that build a 
joint latent space for several modalities. 
We exploit this space for efficient 
retrieval via nearest-neighbor search. %
\PointBind~\cite{guo2023pointBind} and \OpenShape~\cite{liu2023openshape} 
do this 
for text, 3D point clouds (and by extension meshes), and images. 
Here we use \OpenShape to 
efficiently retrieve a likely 3D object mesh~\cite{deitke2023objaverse}
given an image crop around an object; this works even with some occlusion.

\subsection{3D Humans and objects from single images}

Compared to inferring only humans or only objects, jointly inferring them is less explored. 
To support this direction
several datasets have been captured either outdoors~\cite{RICH}
or indoors~\cite{Hassan2019prox, bhatnagar2022behave, huang2022intercap} or 
\camready{have been created} through synthesis \cite{xie2023tprocigen}; but these all consider only constrained settings.
Learned methods \cite{xie2022chore, xie2023tprocigen, xie2023vistracker, nam2024contho, weng2021holistic, shimada2022hulc} 
trained on such data 
either directly regress humans and objects jointly~\cite{xie2023tprocigen, nam2024contho} or 
first infer contact as a proxy and then exploit this proxy for optimization-based fitting~\cite{xie2022chore, xie2023vistracker, shimada2022hulc}.

There has been \camready{significantly} less work addressing in-the-wild settings.
For example,
\PHOSA~\cite{zhang2020phosa}
infers a human mesh with an off-the-shelf model, retrieves an object mesh via mask-based database search, %
and refines its pose 
via hand-crafted category-wise contact constraints. 
Wang~\etal~\cite{wang2022reconstruction} follow a similar strategy, but replace PHOSA's hand-crafted constraints with coarse contact information automatically inferred through an \LLM. %
Both category-wise and LLM-based contacts, however,  lack image-grounding, resulting in inaccurate \hoi reconstruction. In contrast, \nameMethod's inferred contacts consider the input image and generalize to significantly more diverse objects than previous methods. \camready{Moreover, while \PHOSA uses a single scale per class, we use instance-specific scale inferred directly from pixels.}

\subsection{3D Contact estimation}

Studying contact has a long history \cite{kamakura1980grasp, dillmann2005grasp}. 
For example, \obman~\cite{hasson2019obman} generates synthetic grasps~\cite{Miller2004} and uses these to learn likely contacts. 
In contrast, \mbox{ContactDB}~\cite{brahmbhatt2019contactdb} %
captures %
contact regions of real hands grasping 3D-printed objects via thermal imaging, 
while other %
work %
uses alternative 
means %
\cite{lakshmipathy2021contacttracing,yin2022membrane,yin2023extreme}, 
such as marker-based Motion Capture~\cite{GRAB:2020,fan2023arctic}
or multi-view \mbox{RGB-D}~\cite{Brahmbhatt_2020_ECCV}. 
Such work creates %
datasets for 
training methods to 
predict, refine, and associate contacts for pose optimization \mbox{\cite{grady2021contactopt, jiang2021graspTTA, yang2021cpf, zhangNL22024, brahmbhatt2019contactgrasp, lakshmipathy2022contacttransfer, wei2023generalized}}. 
But these datasets are captured in \camready{the} lab, %
so methods trained on them do not generalize. %

\camready{COMA~\cite{kim2024coma} and CHORUS~\cite{han2023chorus} train on synthetic data to predict separate human and object contact distributions, and get rough correspondences via heuristic thresholds on proximity/orientation. 
Instead, 
\nameDataset %
uses 
\emph{fine} manually-set contact correspondences on \emph{real}, natural images.}

More recently, 
\DECO~\cite{tripathi2023deco}, \camready{EgoChoir}~\cite{yang2024egochoir} and \LEMON \cite{yang2023lemon} 
crowd-source contact areas in natural images through online ``vertex painting'' tools. 
\DECO annotators 
``paint'' contact only on the body, while 
\LEMON \camready{and EgoChoir annotate}
on both the 
human and the object in separate processes;
that is, the body and object contacts do not need to correspond. 
We
avoid painting contacts on objects by developing 
a 
novel method that 
projects \DECO's body contacts onto objects 
with minimal human effort. %
Crucially, this also establishes bijective body-object contact correspondences. This goes beyond lab~\cite{RICH, bhatnagar2022behave, huang2022intercap} or synthetic~\mbox{\cite{shimada2022hulc, xie2023tprocigen}} data, or %
part-level contacts~\cite{zhang2020phosa}, 
and serves 3D reconstruction.

\section{\nameDataset Dataset}
\label{sec:dataset}

Training \camready{robust} 3D \hoi methods requires \emph{natural} images paired with \emph{both} 3D human and object contacts. %
The \DAMON dataset~\cite{tripathi2023deco} 
pairs natural images %
\cite{gupta2015vcoco,li2020hake} 
with vertex-level contacts on the \SMPL~\cite{SMPL:2015} body,
but it lacks 3D object shapes and object contact. %

To address this, we build a novel method that retrieves matching 3D object \camready{meshes} %
given in-the-wild images~(\cref{sec:object_shape_retrieval}), and projects \DAMON's body-only contacts onto the retrieved mesh~(\cref{sec:contact_parametrization_transfer}). 
We scale this for crowd-sourcing contact annotations on the internet~(\cref{sec:contact_transfer_interface}). This results in \nameDataset, the first dataset that pairs in-the-wild images with 3D object shapes and contacts on both bodies and objects, as well as correspondences between them. %

\subsection{3D Object shape retrieval}
\label{sec:object_shape_retrieval}

We use \OpenShape~\cite{liu2023openshape}, a model with a joint latent space for images and 3D shapes. 
\camready{Offline}, 
we embed the meshes of the \ObjaverseLVIS~\cite{deitke2023objaverse} database into this space. %
\camready{Online}, 
we embed each test image into 
\camready{this} 
space %
and find the 3 closest object latent codes via cosine-similarity. 
Out of 3 options, an annotator picks the 
\camready{one} 
\camready{best matching} 
the image. 

\subsection{Contact representation \& projection}
\label{sec:contact_parametrization_transfer}

\begin{figure}
    \centering
    \includegraphics[width=0.99 \linewidth]{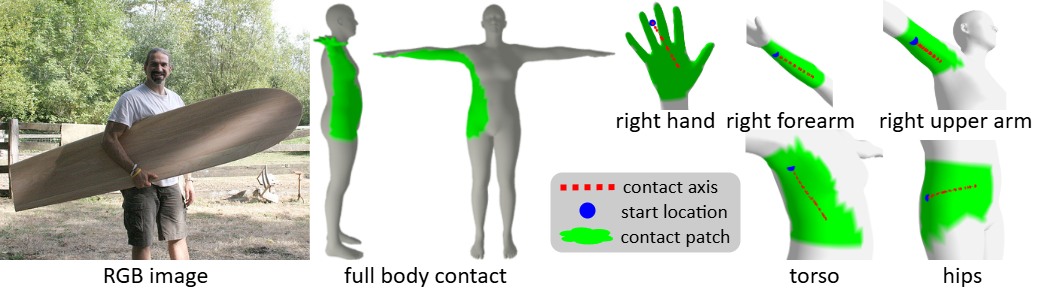}
    \caption{
                Example contact patches with their contact axis. 
    }
    \label{fig:pico:contact_axis}
\end{figure}

\DAMON's body contacts form neighboring-vertex patches. 
We follow ``ContactEdit''~\cite{lakshmipathy2023contactedit} and represent such patches with a contact ``axis'' (see \cref{fig:pico:contact_axis}), \ie an open curve on the patch surface. Since every patch vertex can be parameterized with its contact axis, transferring patches to another surface boils down to transferring only the axis. 
Thus, the axis lets us completely unpack body contact patches %
onto an object with just two clicks, which define the axis start location and direction, respectively. 
Crucially, this also defines bijective point correspondences. For details, see \supmat

However, this approach has two key drawbacks.
First, although ContactEdit infers a default axis, it is originally designed for 3D professionals~\cite{lakshmipathy2023contactedit}, who can ``redraw'' the default axis when it is non-intuitive. This is challenging and time-consuming for non-experts.  
We tackle this by automatically computing a high-quality default axis per patch.
To this end, for each body patch, we perform Principal Component Analysis on its vertex locations. %
Then, starting from the mean, we take
positive and negative steps in the direction of the most significant component, and project the resulting two points onto the body surface via closest point queries~\cite{FCPW}. 
An axis is generated by tracing a geodesic between the two points, while all intermediate triangle edge-crossings serve as axis way-points. 

Second, this approach struggles for contact patches on fine and highly non-convex body areas, such as fingers. 
Specifically, when synthesizing a patch axis, tracing a ``straight" geodesic is hard due to surface concavities. 
Thus, the axis seems 
stretched out when transferred to more planar object regions, 
which confuses %
annotators and also %
distorts patches. 
To tackle this, we create a proxy \smpl mesh with webbed fingers by computing convex hulls for hands; see image in \supmat 
This is a simple, yet effective solution. %

\subsection{Collecting \nameDataset annotations}
\label{sec:contact_transfer_interface}

The contact transfer method of \cref{sec:contact_parametrization_transfer} runs in real time and is user-friendly. 
Thus, we embed this into an interactive web-browser tool. %
For each \DAMON image, we automatically parameterize its 3D body contacts \camready{(via contact axes)} and also retrieve a 3D object shape (see \cref{sec:object_shape_retrieval}). 
Then, we crowd-source annotations for transferring body contacts onto retrieved objects 
via 
Amazon Mechanical Turk. %

\pagebreak

Specifically, for each body-contact patch, annotators click two points on the object mesh -- the first click specifies the start of the contact axis and the second click specifies its %
orientation. 
Then, the tool instantly displays the transferred contact on the object for visual feedback. %
Annotators can correct errors by repeating the two clicks (overwriting past efforts). %
The tool has features such as mesh rotation, zoom in/out, view reset, %
and a menu %
for modifying a previously-annotated contact patch. 
For a detailed visualization and discussion, see \supmat and the \video.

\zheading{\nameDataset statistics}
We annotate 4123 images, %
spanning 44 object categories
and 627 %
object instances. 
To ensure high quality, we select proficient annotators via a qualification process and continuously review their work.
For detailed statistics and quality checks, see \supmat

\section{\nameMethod Method}
\label{sec:pico_fit}

\begin{figure*}
    \vspace{-0.2 em}
    \centering
    \includegraphics[width=0.99\textwidth]{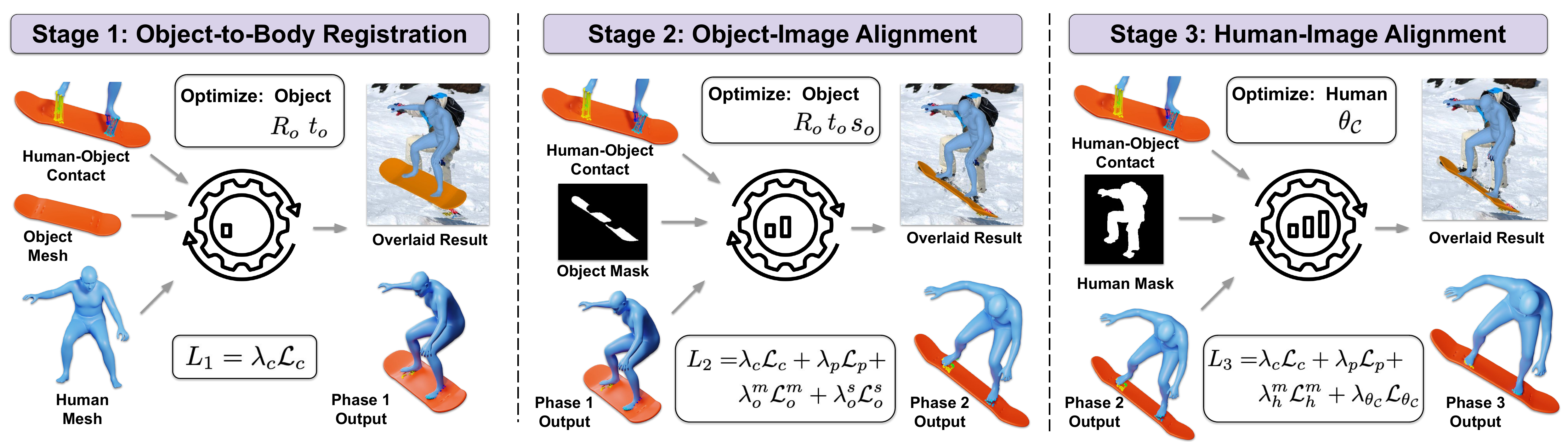}
    \vspace{-0.5 em}
    \caption{
                Overview of \nameMethod, a novel method for fitting interacting 3D body and object meshes to an image. 
                It initializes (\cref{sec:pico_fit_initialization}) 
                3D body shape and pose via \OSX~\cite{lin2023osx}, 
                3D object shape via \OpenShape~\cite{liu2023openshape}, and 
                body-object contacts via %
                retrieval from 
                \nameDataset (\cref{sec:dataset}). 
                Then, it 
                takes three steps: 
                (1)~It exploits contacts to solve for object pose, to register the object to the body (\cref{sec:pico_fit_stage_1}). 
                (2)~It refines object pose (\cref{sec:pico_fit_stage_2}) and (3)~body pose (\cref{sec:pico_fit_stage_3}) 
                to align these to an object and human mask, respectively, detected in the image
                while satisfying contacts and avoiding penetrations. 
                For every stage we show 
                inputs, outputs, losses, and optimizable variables. 
                \faSearch~\textbf{Zoom in} to see details.
    }
    \label{fig:pico:optimization_pipeline}
    \vspace{-0.4 em}
\end{figure*}

We develop \nameMethod, a novel method that, given an image 
$I$, 
recovers a 3D human and object mesh 
realistically registered \wrt each other. %
Learning this is intractable due to the lack of 3D \HOI datasets. 
\camready{Thus, we leverage contact correspondences between the body and the object to fit 3D meshes to images.}
But this is hard due to strong occlusions and depth ambiguities. 
We tackle these via a careful initialization and three-stage fitting (see \cref{fig:pico:optimization_pipeline}) as follows.

\subsection{Initialization}
\label{sec:pico_fit_initialization}

\zheading{Body shape \& pose initialization}
We apply the \OSX~\cite{lin2023osx} regressor on image $I$ to infer 
a \smplX \cite{SMPL-X:2019} body mesh, 
$\humanmesh$, with initial pose $\theta^*$, 
that has articulated hands. %

\zheading{Object shape initialization}
We \camready{apply} the method of \cref{sec:object_shape_retrieval}, \ie,
we use the \OpenShape~\cite{liu2023openshape} model that embeds images and 3D shapes into a single latent space, $\mathcal{G}$. 
Offline, we embed into $\mathcal{G}$ the \ObjaverseLVIS~\cite{deitke2023objaverse} meshes; 
for each object mesh $\objectmesh_i$, we get a latent code $g_i$. 
Then, we embed image $I$ to get the latent code $g_{in}$ and find the closest code to it, 
$g_j = \text{argmax}_j \frac{g_j \cdot g_{in}}{\|g_j\| \|g_{in}\|}$, 
encoding the object mesh $\objectmesh_j$ that best matches the image. 
This is automatic, fast, robust to \camready{some} occlusions, preserves 3D details, scales well, %
and easily handles new object classes as databases get richer. %
We initialize scale, $s^*_\object$, via \GPT; see details in \supmat

\zheading{Contact initialization}
For 
\nameDataset images, 
we %
use %
the associated annotations. %
\camready{For unlabeled images, we need to infer contact correspondences. 
However, 
no method can infer %
contact correspondences on the object \wrt the body, 
due to 
the huge object shape variance. 
Our key insight is to exploit 3D contact on \camready{bodies}, %
which is easier to infer, %
as 
key 
to ``query'' %
the respective object contact from \nameDataset. }

\camready{To this end, we infer vertex-level body contact via \DECO \cite{tripathi2023deco}}. 
But this is often noisy, as this problem is unsolved. 
Thus, we further ask \GPT \emph{``which $\langle$body part$\rangle$ is in contact with the~$\langle$object$\rangle$''} 
to 
reduce false negatives 
(in general), 
and 
false positives on feet 
\camready{(\DECO's bias)}; 
for details see \supmat 
\camready{The} estimated contact helps ``query'' \nameDataset to retrieve %
the closest body-contact annotation that maximizes the intersection-over-union (IoU) between these. 
This is inspired by seminal work~\cite{weyand2016nnsearchVSnetworks} showing that 
nearest-neighbor retrieval from a rich database can be better than regression. 

Since \nameDataset body contact is paired with 3D object shape, object contact, and body-object contact correspondences, we also retrieve these for free to initialize \nameMethod. 

\subsection{Stage 1: Register\camready{ing} object to body via contact}
\label{sec:pico_fit_stage_1}

At this point, 
we have initialized 3D body shape and pose, 3D object shape, and body-object contacts. 
However, the object pose remains unknown. 
To tackle this, we keep the human fixed and use contact correspondences to solve for object pose, \ie, rotation, $R_\object\in\mathbb{R}^{3}$, and translation, $t_\object\in\mathbb{R}^{3}$. 
In detail, we use body-to-object (vertex-to-point) correspondences %
$\mathbb{S} := \{ (v_i, p_i) \}$
where 
$v_i\in\humanmesh$ are human-mesh vertices, while $p_i\in\objectmesh$  %
are points (that might lie inside triangles) on the object surface. 
Then, we 
estimate $R_\object$ and $t_\object$ %
and register the object to the body 
by minimizing a contact loss: 
$\loss_1 = 
\loss_\contact = 
\frac{1}{|\contactset|}\sum\nolimits_{(v_i, p_i)\in\contactset}\| v_i \; - \; p_i \|_2$. 

However, all regressors are imperfect, so \OSX-inferred bodies can be noisy, especially for challenging images. 
This also affects object pose. 
So, after stage~1, human and object meshes might be image-misaligned and need refinement. 
To avoid chicken-and-egg problems in joint refinement, we first refine the object \mbox{(stage~2)} and then the body \mbox{(stage~3)}. 
Empirically, an extra joint-refinement stage does not help.

\subsection{Stage 2: Aligning the object to the image} %
\label{sec:pico_fit_stage_2}

Here we refine the object to align with the image. %
First, 
we render the object mesh, $\objectmesh$, %
into a 2D object mask, $\predobjmask$, 
via an \OSX-inferred %
camera  %
and the \mbox{PyTorch3D} renderer. 
Then, we  
detect in image $I$ an object mask, $\gtobjmask$, via %
\SAM \cite{ren2024groundedsam}. 
Last, with \IoU($\cdot$) denoting Intersection-over-Union, 
we define the object mask loss: 
$\loss_\object^\mask 
= 1 - \text{IoU}(\gtobjmask, \predobjmask)$.

However, 
this might cause human-object penetrations, %
as it ignores %
the relative 3D human-object arrangement. 
We tackle this here. %
Let $\mathbf{\Omega}_\human$ be a %
Signed Distance Field (SDF)~\cite{Hassan2019prox} %
around the human mesh, $\humanmesh$. 
For a 3D point, $\mathbf{\Omega}_\human$ has values 
proportional to the distance from $\humanmesh$, %
with a positive sign for points inside $\humanmesh$ and negative outside $\humanmesh$. 
Then, the penetration loss~\cite{jiang2020multiperson}, 
$\loss_\collision = \sum_{
v_i 
\in \objectmesh} \mathbf{\Omega}_\human(
v_i)$,
runs over all object vertices, 
$v_i$, 
paying a penalty, $\mathbf{\Omega}_\human(x,y,z) = -\text{min}(\text{SDF}(x,y,z), 0)$, 
when $v_i$ penetrates the body. 
Note that related work~\cite{SMPL-X:2019, zhang2020phosa, xie2022chore} penalizes only shallow~\cite{ballan2012collision} penetrations and misses extreme ones, in contrast to our $\loss_\collision$.  

We also define an object scale loss, 
$\loss_\object^\scale = 
\| s_\object \; - \; s^*_\object \|_2$, to refine the scale based on image evidence (see $\loss_\object^\mask$ above) while not deviating much from the initial estimate, $s^*_\object$. 

We optimize over object \camready{rotation}, %
$R_\object$, translation, $t_\object$, and scale, $s_\object\in\mathbb{R}$. 
With $\lambda$ denoting steering weights tuned empirically, we minimize:
$L_2 = 
\lambda_\contact        \loss_\contact      + 
\lambda_\collision      \loss_\collision    + 
\lambda_\object^\mask   \loss_\object^\mask + 
\lambda_\object^\scale  \loss_\object^\scale$.

\subsection{Stage 3: \camready{Refining the human pose}}
\label{sec:pico_fit_stage_3}

\camready{The goal %
is to refine the contact between the human and the pixel-aligned object from Stage 2. 
To this end, %
we employ the contact loss $\loss_\contact$ to optimize the 
human 
pose.} %
\camready{However, this loss alone %
does not provide enough constraints and 
may lead to implausible poses.}
\camready{Thus, we add two regularizers.} %

\camready{
First, we define a human mask loss, like the object one in Stage 2.
Using the same camera as for objects, we render the human mesh, $\humanmesh$, as a 2D mask, $\predhumanmask$. 
We also detect in image $I$ a human mask, $\gthumanmask$, via \SAM~\cite{ren2024groundedsam}. 
Then, with \IoU($\cdot$) denoting Intersection-over-Union, the mask loss is 
$\loss_\human^\mask = 1 - \text{IoU}(\gthumanmask, \predhumanmask)$. 
}
But minimizing $\loss_\human^\mask$ by 
optimizing over $\theta$ 
produces distorted bodies due to depth ambiguity. 
To tackle this we need another regularizer, %
$\loss_r = \| \theta-\theta^* \|_2$, 
so that pose $\theta$ does not deviate much from the initial $\theta^*$. 

Interestingly, we observe that the initial body has a good torso pose, but errors increase towards end effectors. 
Thus, we optimize only the pose parameters for the limbs after the torso until the ones contacting the object. 
Assuming just one contacting limb for notational simplicity, let 
$\mathcal{C} = \{ J_r, J_{r+1}, \dots, J_\contact \}$ be the %
joints 
from the closest torso joint, $J_r$, 
to the contacting joint, $J_\contact$, 
along the kinematic chain. 
Then, we only optimize over %
$\theta_\mathcal{C} = 
\{ \theta_r, \theta_{r+1}, \dots, \theta_\contact \}$. 
With $\lambda$ denoting steering weights tuned empirically, we minimize:
$L_3 = 
\lambda_\contact        \loss_\contact      + 
\lambda_\collision      \loss_\collision    + 
\lambda_\human^\mask    \loss_\human^\mask  + 
\lambda_{\theta_\mathcal{C}}   \loss_{\theta_\mathcal{C}}$.

\section{Experiments}
\label{sec:experiments}

Existing 3D \hoi recovery methods \cite{nam2024contho, xie2022chore, xie2023tprocigen} 
perform well 
on datasets they train on. 
However, they fail for %
out-of-domain (OOD) scenarios, \ie: (1) unseen in-lab datasets, and (2) unseen %
\inthewild images; 
the latter is the main focus of our work. 
Thus, we compare our \nameMethod method with both regression-based \hoi reconstruction methods, \ie, \CONTHO~\cite{nam2024contho} and \HDM~\cite{xie2023tprocigen}, and an optimization-based one, \ie, \PHOSA~\cite{zhang2020phosa}, on these tasks. 

\zheading{Datasets}
To evaluate 3D \hoi reconstruction, two in-lab datasets are widely used, \InterCap~\cite{huang2022intercap} and \BEHAVE~\cite{bhatnagar2022behave}. 
These %
provide multi-view \mbox{RGB-D} images paired with 3D ground-truth (GT) %
bodies and objects in interaction, with 10 and 20 %
objects, respectively. 
Most %
methods train on \BEHAVE. %
So, we use \BEHAVE-trained checkpoints for existing methods, and evaluate %
generalization to \InterCap. 

\zheading{Metrics}
Following past work~\cite{xie2022chore, nam2024contho, xie2023vistracker}, we report the Procrustes-Aligned (PA) Chamfer Distance (CD). 
We compute this separately on the \smplX body \mbox{(PA-CD$_\human$)} and object mesh \mbox{(PA-CD$_\object$)},
after \camready{first} performing PA to align the combined \camready{human+object} mesh with the GT meshes. 

\subsection{3D \HOI reconstruction -- OOD in-lab datasets}
\label{sec:experiment:ood-inlab-datasets}

\begin{table}[t]
    \centering
    \resizebox{\columnwidth}{!}{
        \begin{tabular}{lccc|ccc|c}
            \toprule
            \multirow{3}{*}{\bf Methods} & \multirow{3}{*}{Ref.} & \multirow{3}{*}{Type} & \multirow{3}{*}{\centering \shortstack{GT\\con-\\tact}} & \multicolumn{3}{c|}{\InterCap~\cite{huang2022intercap}} & \DAMON~\cite{tripathi2023deco} \\
            \cmidrule{5-7}\cmidrule{7-8}
               &  & & & \bf PA-CD$_\human$ & \bf PA-CD$_\object$ & \bf PA-CD$_{\human+\object}$ & X vs \textbf{\nameMethod}$^*$  \\
               &  & & & (cm) $\downarrow$  & (cm) $\downarrow$ & (cm) $\downarrow$ & Pref. Rate (\%) \\ %
            \midrule
            \HDM    & \cite{xie2023tprocigen} & Reg. & \textcolor{red}{\xmark} & 17.34 & 14.12 & 13.6 & 20.1 vs 79.9 \\
            \CONTHO & \cite{nam2024contho}    & Reg. & \textcolor{red}{\xmark} & 8.36 & 24.30 & 13.14 & - \\
            \PHOSA  & \cite{zhang2020phosa}   & Opt. & \textcolor{red}{\xmark} & 10.07 & 23.36 & 13.38 & -  \\
            \midrule
            $\text{\CONTHO}^*$  & \cite{nam2024contho}  & Reg. & \textcolor{green}{\cmark} & 8.16 & 23.26 & 12.81 & 24.7 vs 75.3 \\
            $\text{\PHOSA}^*$   & \cite{zhang2020phosa} & Opt. & \textcolor{green}{\cmark} & 10.12 & 20.91 & 13.28 & 32.0 vs 68.0 \\
            \midrule
            $\textbf{\nameMethod}$   & Ours & Opt. & \textcolor{red}{\xmark} & 7.43 & 21.85 & 10.33 & 37.3 vs 62.7 \\
            $\textbf{\nameMethod}^*$ & Ours & Opt. & \textcolor{green}{\cmark} & \textbf{6.66} & \textbf{13.34} & \textbf{8.36} & $\varnothing$ \\

            \bottomrule
        \end{tabular}
    }
    \vspace{-0.3 em}
    \caption{
                Evaluation on 3D \HOI reconstruction. 
                \mbox{\qheading{Middle column}} 
                Evaluation on \InterCap~\cite{huang2022intercap} (\cref{sec:experiment:ood-inlab-datasets}). 
                Since no method trains on \InterCap, this evaluates generalization. %
                \mbox{\qheading{Right column}} 
                Evaluation on \inthewild images via a perceptual study (\cref{sec:experiment:ood-inthewild}). 
                We report the preference rate of results from the competing method (denoted as ``X'') 
                over our \textbf{\nameMethod}$^*$. 
                \mbox{\qheading{Left column}} 
                 ``Type''  denotes regression %
                 or optimization. %
                Using GT contact is %
                highlighted with $^*$. %
    }
    \label{tab:intercap_quant_results}
\vspace{-0.4 em}
\end{table}

We evaluate 
\nameMethod 
and 
\SOTA methods 
on \InterCap \cite{huang2022intercap} and report results in \cref{tab:intercap_quant_results}. 
The \CONTHO model trains %
on \BEHAVE, 
\camready{while \HDM trains on %
\Procigen \cite{xie2023tprocigen}, a %
synthetic dataset 
building on \BEHAVE and \InterCap. 
So, with 
the %
exception of \HDM, 
\InterCap is unseen for all models.}
Thus, we evaluate generalizability for OOD in-lab images. 

Further, %
we ablate the impact of using ground-truth (GT) 3D contacts 
extracted from GT human and object \InterCap meshes %
with a distance threshold of 5 cm. %
This simulates perfect contact ``detection'' to provide an upper bound on accuracy. %
Methods that use GT contact are highlighted with a star~($^*$); %
see details for each method in \supmat

When GT contact is available, \nameMethodStar significantly outperforms all baselines. However, even \nameMethod, 
\camready{which does not use} 
GT contact, performs on par with  $\text{\PHOSA}^*$ and $\text{\CONTHO}^*$, demonstrating its robustness.

\subsection{3D \HOI reconstruction -- In-the-wild images}
\label{sec:experiment:ood-inthewild}

We evaluate \nameMethodStar against \SOTA methods on \inthewild images %
through a perceptual study conducted on Amazon Mechanical Turk. %
We randomly select 75 images from 42 object categories in the \DAMON dataset, and evaluate each method on these samples. 

Participants are shown an image at the center, along with reconstructions from \nameMethodStar and baselines, randomly shuffled to the left and right side. 
The participants mark which of the two reconstructions best reflects the image, 
while focusing on the %
3D human-object contact and spatial alignment. 
For details about the study, see \supmat %

Note that 
$\text{\CONTHO}^*$ is only trained on 9 object classes. %
To ensure fair comparison, we evaluate $\text{\CONTHO}^*$ on 30 images that span only these specific 9 objects. 
Note also that \HDM outputs a point cloud, while our \nameMethodStar produces meshes. 
To avoid introducing any visualization bias, we convert \nameMethodStar meshes to point clouds. %

We report results in \cref{tab:intercap_quant_results} (right), 
in the form of ``X vs \nameMethodStar'', indicating the percentage of times a competing method (``X'') was preferred over our method. %
On average, participants deemed reconstructions produced by \nameMethodStar to be more realistic over baselines 74.4\% of the time.

\begin{figure}
    \centering
    \includegraphics[width=0.99 \linewidth]{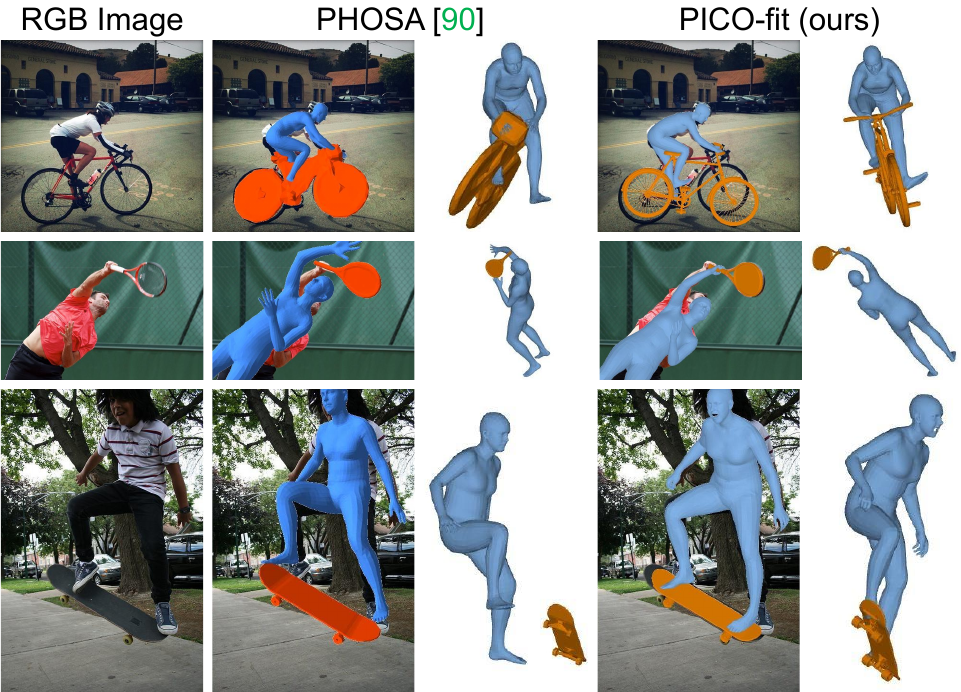}
    \caption{
                Qualitative comparison of \nameMethod vs \PHOSA on internet images used for evaluation in the \PHOSA paper \cite{zhang2020phosa}. 
    }
    \label{fig:experiments:phosa_comparison}
\end{figure}

\begin{figure*}
    \centering
    \includegraphics[
    width=\linewidth]{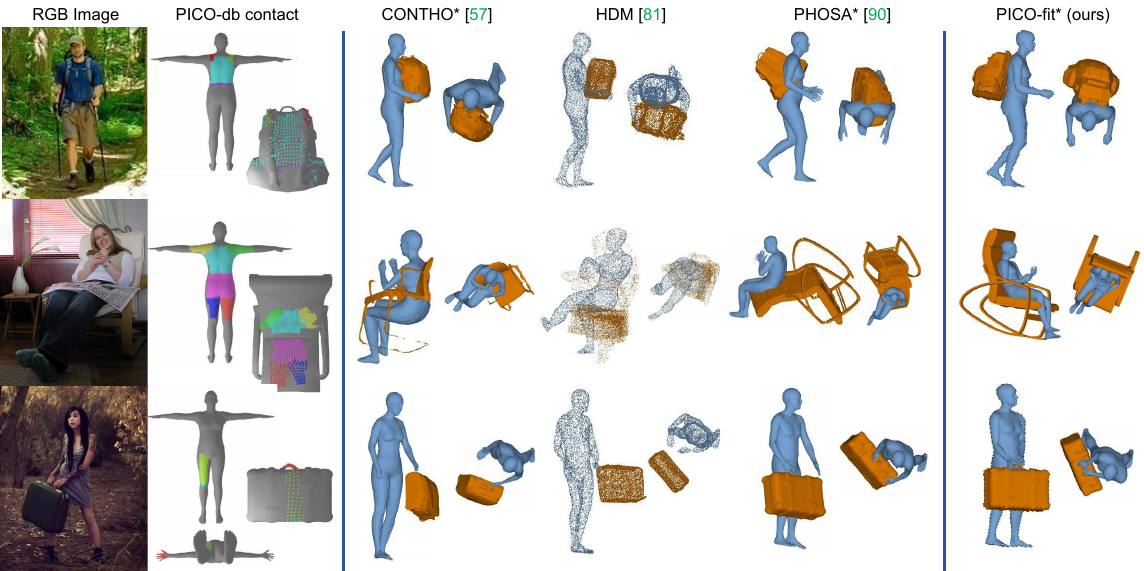}
    \caption{
        Qualitative evaluation of $\text{\CONTHO}^*$, \HDM and $\text{\PHOSA}^*$ alongside \nameMethodStar on object categories handled by all baselines. 
        \camready{From left to right: 
        input image, pseudo-GT contact annotations in \nameDataset, and 3D reconstructions (a side and top-down view per method)}. 
        Reconstructions from \nameMethodStar have better 3D human-object contact and spatial alignment. For more comparisons, see \supmat
    }
    \label{fig:experiments:pico_sota_comparison}
    \vspace{+1.0 em}
\end{figure*}

\begin{figure*}
    \centering
    \includegraphics[trim=000mm 000mm 000mm 000mm, clip=true, width=\linewidth]{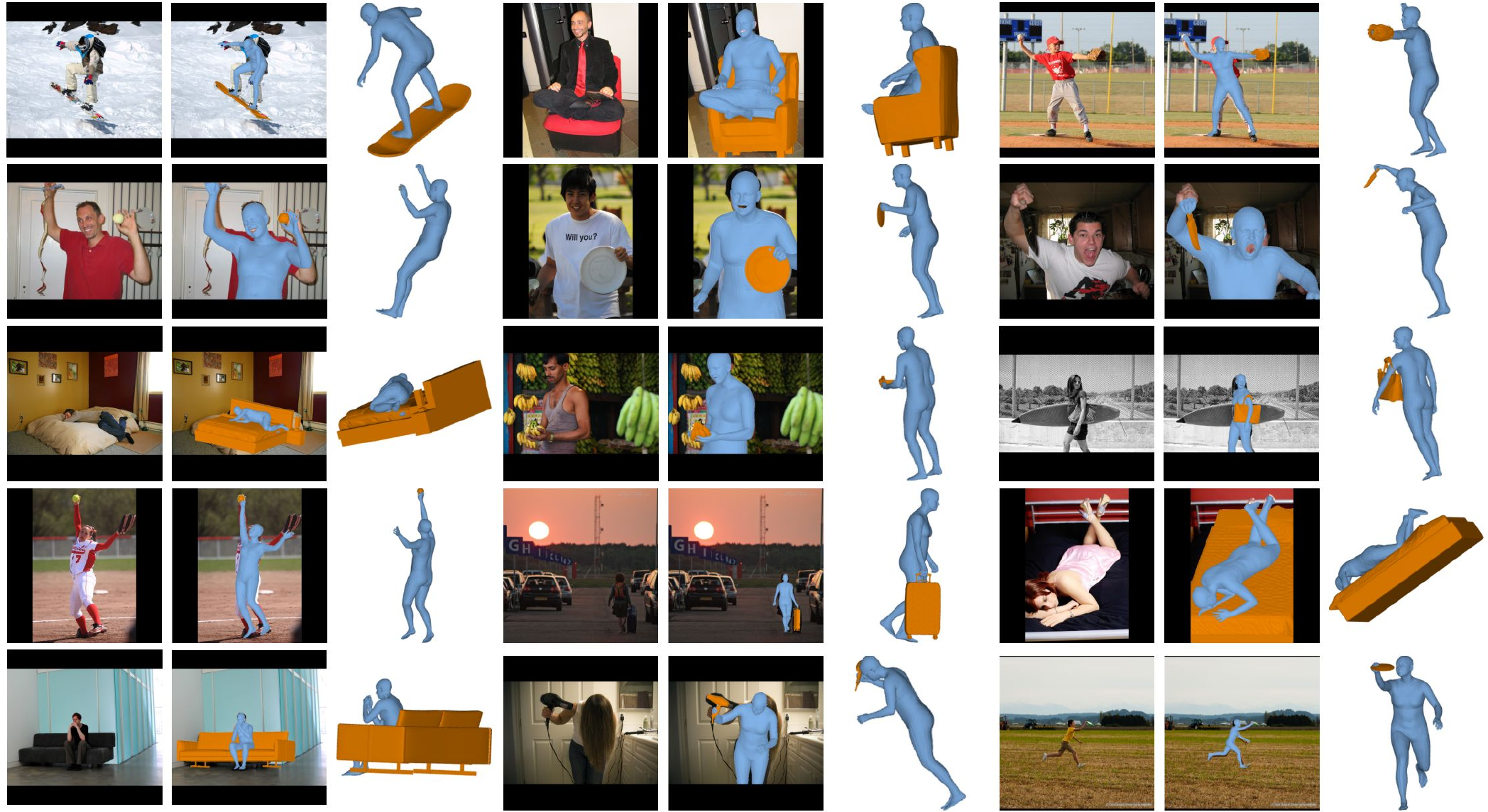}
    \caption{\hoi reconstructions from \nameMethodStar on \emph{new, previously untackled} object categories. Each row (left to right) shows, for three input RGB images, \nameMethodStar's estimated meshes overlaid on the image (camera view) and a side view. For more results, see \supmat}
    \label{fig:experiments:pico_inthewild_1}
\end{figure*}

\pagebreak

\zheading{Qualitative evaluation} 
In \cref{fig:experiments:pico_sota_comparison} we qualitatively compare \nameMethodStar with \SOTA methods on \DAMON images, only for object categories %
handled by all baselines. 
In \cref{fig:experiments:pico_inthewild_1} we show \nameMethodStar reconstructions on object categories that no previous method can handle. 
Finally, 
in \cref{fig:experiments:phosa_comparison}, we qualitatively compare \nameMethod with \PHOSA, namely the most related \SOTA method to ours, on the same internet images used in \camready{the \PHOSA paper \cite{zhang2020phosa}}. 
We show qualitative comparisons of \nameMethod with other baselines in \supmat %

These results show that \nameMethod is more robust and generalizes to challenging natural images better than existing methods. 
Note that \nameMethod handles several object classes for the first time, 
due to  efficient 
retrieval from \nameDataset. %

\subsection{Ablation study}

\begin{table}
\centering
\resizebox{1.00 \columnwidth}{!}{
\begin{tabular}{l|cccc|l|ccc}
\Xhline{3\arrayrulewidth}
\multicolumn{1}{c|}{\bf Stage} & \multicolumn{4}{c|}{\bf Losses} & \multicolumn{1}{c|}{\bf Optimized} & \multicolumn{3}{c}{\bf Procrustes-Aligned (PA)}\\ 
\multicolumn{1}{c|}{\bf IDs} & $\boldsymbol{\mathcal{L}_\contact}$ & $\boldsymbol{\mathcal{L}_{\object,\mask}}$ & $\boldsymbol{\mathcal{L}_\collision}$ & $\boldsymbol{\mathcal{L}_{\human,\mask}}$ & {\bf Variables} & \bf CD$_\human$ $\downarrow$ & \bf CD$_\object$ $\downarrow$ & \bf CD$_{\human+\object}$ $\downarrow$ ~\\ 

\cline{1-9}

1     & \cmark & \xmark & \xmark & \xmark & $R_\object, t_\object$            & 7.25 & 24.51 & 11.47 \\ \hline
1+2   & \cmark & \cmark & \cmark & \xmark & $R_\object, t_\object, s_\object$ & 6.65 & 13.67 & 8.40  \\ \hline
1+2+3 (\nameMethod) & \cmark & \cmark & \cmark & \cmark & $R_\object, t_\object, s_\object, \theta_\mathcal{C}$ & 6.66 & 13.34 & 8.36  \\ \hline

\end{tabular}
}
\vspace{-0.5 em}
\caption{
            Ablation study for 
            \nameMethod's three fitting stages. 
            We evaluate on the \InterCap~\cite{huang2022intercap} dataset, and report the Procrustes-Aligned Chamfer Distance (PA-CD) for the human ($\human$), object ($\object$), and their combination ($\human+\object$). 
            The middle columns show the %
            losses and optimized variables. 
            For qualitative ablation, see \supmat 
}
\label{table:pico:model_ablation}
\vspace{-0.6 em}
\end{table}

\camready{We evaluate the contribution of \nameMethod's stages %
in \cref{table:pico:model_ablation}. 
This shows 
that each stage contributes meaningfully, as the accuracy significantly improves for the optimized elements, while non-optimized ones either improve or, in the worst-case, remain practically unchanged.
For quantitative ablations on alternate optimization strategies and qualitative ablations
on the effect of each \nameMethod stage, %
see \supmat}

\section{Conclusions and Future Work} %
Our work emphasizes how contact, on both the human body and the objects it interacts with, is a foundation for reasoning about 3D HOI. 
Specifically, we build
a new dataset that uniquely pairs natural images with 3D contacts on both the body and the object. 
Using this, we develop a novel method that exploits contacts to reconstruct 3D \HOI from a single image. 
Our method handles object classes that no existing method handles, 
\camready{via} 
efficient retrieval from our rich \camready{dataset}. %

The next step is to make 3D contact estimation more general, efficient, and robust. 
To that end, we plan to expand and leverage our dataset of \inthewild contact labels to train a direct %
contact regressor. 
Specifically, we will leverage \nameMethod to automate the creation of pseudo ground truth training labels.
With sufficient training data we should be able to replace our nearest-neighbor lookup from \nameDataset with a feed-forward model. 
\camready{Last, we will explore vision-language models \cite{dwivedi2025interactVLM} to go beyond finite datasets.}

\pagebreak

{\small
\qheading{Acknowledgements} 
\camready{We thank 
Felix Grüninger for advice on mesh preprocessing, 
Jean-Claude Passy and Valkyrie Felso for advice on the data collection, and 
Xianghui Xie for advice on HDM evaluation. We also thank Tsvetelina Alexiadis, Taylor Obersat, Claudia Gallatz, Asuka Bertler, Arina Kuznetcova, Suraj Bhor, Tithi Rakshit, Tomasz Niewiadomski, Valerian Fourel and Florentin Doll for their immense help in the data collection and verification process, Benjamin Pellkofer for IT support,
and Nikos Athanasiou for the helpful discussions.
This work was funded in part by the International Max Planck Research
School for Intelligent Systems (IMPRS-IS).   
D. Tzionas is supported by the ERC Starting Grant (project \mbox{STRIPES}, \mbox{101165317})}}

\noindent
{\small \textbf{{Disclosure}:} 
\camready{DT has received a research gift fund from Google. 
For MJB see \href{https://files.is.tue.mpg.de/black/CoI\_CVPR\_2025.txt}{https://files.is.tue.mpg.de/black/CoI\_CVPR\_2025.txt}}}

{
    \small
    \bibliographystyle{config/ieeenat_fullname}
    \bibliography{config/BIB}
}

\clearpage
\pagebreak
\maketitlesupplementary

\renewcommand{\thepage}{S.\arabic{page}}
\renewcommand{\thesection}{S.\arabic{section}}
\renewcommand{\thefigure}{S.\arabic{figure}}
\renewcommand{\thetable}{S.\arabic{table}}
\renewcommand{\theequation}{S.\arabic{equation}}
\setcounter{page}{1}
\setcounter{section}{0}
\setcounter{figure}{0}
\setcounter{table}{0}
\setcounter{equation}{0}

\noindent 
Here we provide a detailed description of the \nameDataset data collection~(\cref{supmat:pico_db_details}), including implementation details, data statistics and quality control. In \cref{supmat:vllm_contact_estimation}, we discuss how we leverage vision large language models (VLLMs) to augment body contact prediction from \DECO. In \cref{supmat:evaluation_details}, we provide implementation details about our \nameMethod reconstruction method, along with its quantitative evaluations and perceptual study. Finally, in \cref{supmat:pico_fit_extra}, we provide \camready{additional ablations}, qualitative results and failure cases. %

\noindent \textbf{\Video.} To crowd-source 3D contact annotation on both the body and the object using
Amazon Mechanical Turk (AMT), we build a new annotation
tool which we describe in detail in the following section. We recommend that readers view the provided supplemental video for an in-depth tour of our tool, its features and the annotation protocol. This is the same video we used for training AMT workers before qualifying them for this task.

\section{\nameDataset data collection}
\label{supmat:pico_db_details}

\subsection{Contact representation \& projection details}

\DAMON's body contacts form neighboring-vertex patches. 
To represent such a patch, we compute the axis-based parameterization 
of ``ContactEdit''~\cite{lakshmipathy2023contactedit} 
via 
three steps:
(1)~We synthesize an ``axis,'' \ie, an 
open curve on the body composed of piece-wise shortest geodesics between 
constituent surface points.
(2)~For each patch 
vertex, 
we compute its closest axis point via short-time heat diffusion~\cite{sharp2019vhm},   
and (3)~its logarithmic map ({\tt{logmap}}), \ie, its 
geodesic distance and direction 
\wrt its associated axis point~\cite{MMP:1987}.

The {\tt{logmap}} 
helps 
transferring patches across meshes. %
That is, given an axis, we can reconstruct the patch via the inverse operation, the exponential map ({\tt{expmap}}). 
Thus, transferring patches boils down to transferring only the axis. 
The axis 
can be completely unpacked on any surface given only the starting location and direction of the first geodesic. 

\begin{figure}
    \centering
    \includegraphics[width=1.0 \linewidth]{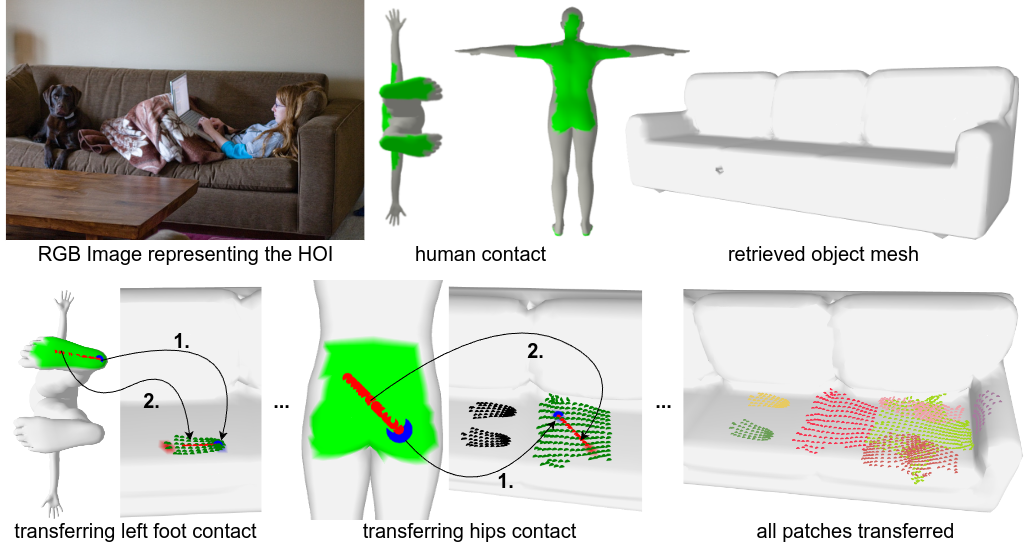}
    \caption{
                \textbf{Top row:} 
                Inputs to the \nameDataset app.
                (1) RGB image.
                (2) \DAMON body contacts.
                (3) The retrieved object mesh. 
                \textbf{Bottom row:} 
                Contact transfer via 2 clicks from the foot (left) and hips (center) onto the object. 
                The first click specifies the axis start location (blue ball), and the second one the axis direction (red line). 
                \textbf{Bottom row (right):}
                Resulting patches after annotation.
    } 
    \label{fig:supmat:contact_transfer_steps}
\end{figure}

Simply put, this lets us transfer body contact patches %
onto an object with just two clicks, which define the axis start location and direction respectively. 
Crucially, this also defines bijective point correspondences 
between patches\footnote{Formally, given a source and target patch, there exists a point mapping
that is theoretically-guaranteed 
surjective, but empirically it is bijective~\cite{lakshmipathy2023contactedit}.}. 
The axis parameterisation enables the automatic correspondence of discretised contact areas, comprising hundreds or even thousands of points, between the body and the object through an intuitive and substantially lower-dimensional representation. 
\Cref{fig:supmat:contact_transfer_steps} and Fig.~\textcolor{red}{3} in main illustrates several body contacts and their respective axes. For further details, we refer the reader to Lakshmipathy \etal~\cite{lakshmipathy2023contactedit}.

\subsection{Projection using the proxy mesh}

\begin{figure}
    \centering
    \includegraphics[width=0.99 \linewidth]{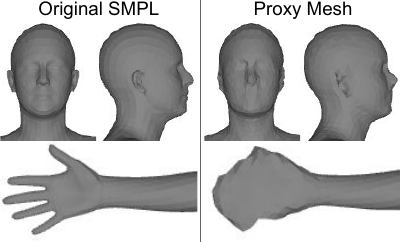}
    \vspace{-0.5em}
    \caption{Comparison of the face and hand details between the original \smpl human body mesh and the proxy mesh with ``webbed" hands and ``smoothed-out'' face features, used for simplifying contact patch projections.}
    \label{fig:supmat:smpl_simplified_mesh}
\end{figure}

Because transferring patches parameterized on non-convex shape regions can yield non-intuitive results, we construct a proxy \smpl mesh to ``convexify" the hands and face features. Specifically, we take a convex hull of the hands and improve the triangulation via tangential smoothing~\cite{chen2011smoothing} and Delaunay refinement~\cite{shewchuck2002delaunay} using the Geometry Central library~\cite{geometrycentral}. The result (\cref{fig:supmat:smpl_simplified_mesh}) is a \smpl mesh with ``webbed" hands and ``smoothed-out'' face features. 

We first project \DECO contacts from the original \smpl body to the proxy ``convexified'' body via closest point queries~\cite{FCPW}. 
We then parameterize contacts on the proxy body, and last transfer these to objects. 
However, for visualization purposes, we present annotators with contacts on the original body with overlaid axes from the proxy body.

Note that these ``convexified'' meshes are used only as a ``proxy tool'' to ease defining ``straight'' contact axes. 
We do not use these later for 3D reconstruction, so the accuracy of reconstructions is not compromised. 

\subsection{Object mesh processing}

We rely on the Objaverse-LVIS~\cite{deitke2023objaverse} dataset for retrieving object meshes. However, our contact parameterization and projection requires input meshes to be \emph{manifold}, which is not true for several meshes in \mbox{Objaverse-LVIS}. Therefore, we perform a series of pre-processing operations to curate a database of manifold objects. Specifically, we use the \mbox{PyMeshLab} library~\cite{pymeshlab} and apply Poisson-Disk sampling to generate 50k uniformly sampled surface points. Next, we perform Screened Poisson surface reconstruction and uniform mesh resampling on the resulting point cloud, with parameters: depth $=8$ and samples-per-node $=8$. We also remove floating isolated connected-component pieces which do not belong to the original mesh. This produces a smooth mesh, which is further corrected using the ``3D Print Toolbox''~\cite{3dprintblender} in Blender to ensure manifoldness. Last, we verify manifoldness and discard any non-manifold objects.

Although the processed meshes are manifold, they are often too high-resolution %
and arbitrarily scaled, making them unsuitable for online operations in our data annotation app. To address this, we decimate the meshes to a maximum of 4k faces using Blender. Additionally, we recenter the meshes to the origin and rescale them by querying GPT-4V to determine the correct scale based on the corresponding RGB image input.

\subsection{Contact annotation tool}

Following \cite{tripathi2023deco}, we build the tool in DASH and deploy it inside a Docker container under an uWSGI application server. 

\begin{figure*}
    \centering
    \vspace{-0.5 em}
    \includegraphics[width=1.0 \linewidth]{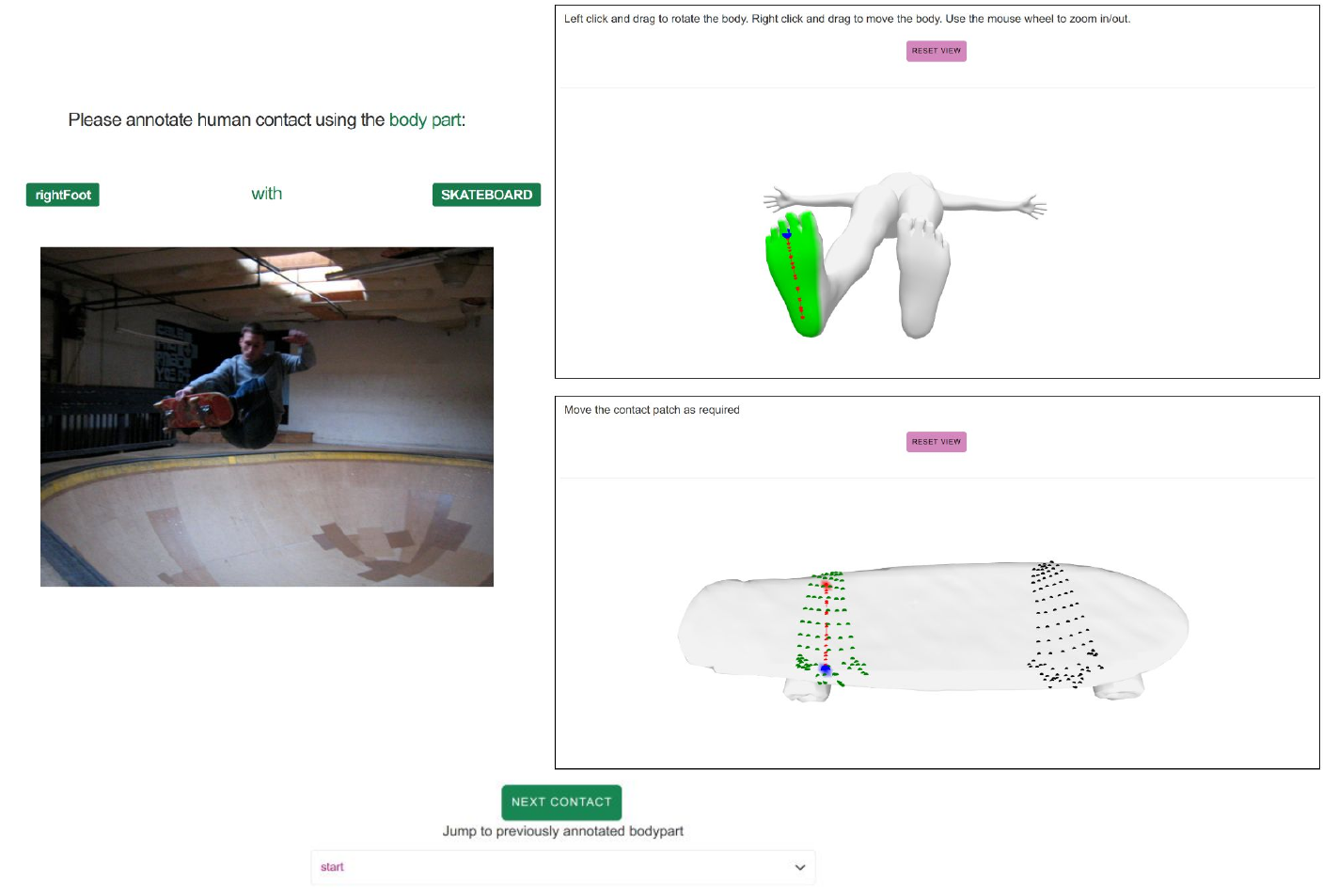}
    \vspace{-1.5 em}
    \caption{
                    Layout of the contact annotation tool. 
                    \textbf{Left side:} 
                    Original image with labels for the object category and the current body part above it. 
                    \textbf{Right side:} 
                    Human mesh in \mbox{T-pose} with contact regions (shown in \textcolor{green}{green} color) and contact axis (shown in \textcolor{red}{red} color). 
                    Below that, we show the 3D object, with the left foot contact already transferred and the right foot contact being annotated in the current step. 
                    Users can navigate back to previous steps (body parts) with the help of the drop-down menu situated below the ``Next contact'' button. 
    }
    \label{fig:supmat:contact_transfer_app}
\end{figure*}

\zheading{Annotation interface} As shown in the tool interface in \cref{fig:supmat:contact_transfer_app}, the layout is divided into 3 parts. On the left, we show annotators the original image with the ``transfer'' candidates denoted as tuples: {\tt \{body part name, object label\}}. On the top-right, we show the human \mbox{T-pose} mesh with \DAMON's contact regions (shown in \textcolor{green}{green} color) and contact axis (shown in \textcolor{red}{red} color). On the bottom-right, we show the 3D object. 

The annotators are required to click two points on the object mesh -- the first click specifies the start of the contact axis and the second click specifies its geodesic orientation. Upon registering the second click, the tool instantaneously displays the transferred contact on the object, providing visual feedback in real time. Annotators can correct errors by repeating the %
two clicks, which resets the prior annotations, until satisfied. For a detailed overview of the tool and its functions, please watch the %
\video.

\subsection{\nameDataset additional statistics}

\begin{figure*}
    \centering
    \includegraphics[width=1.0 \linewidth]{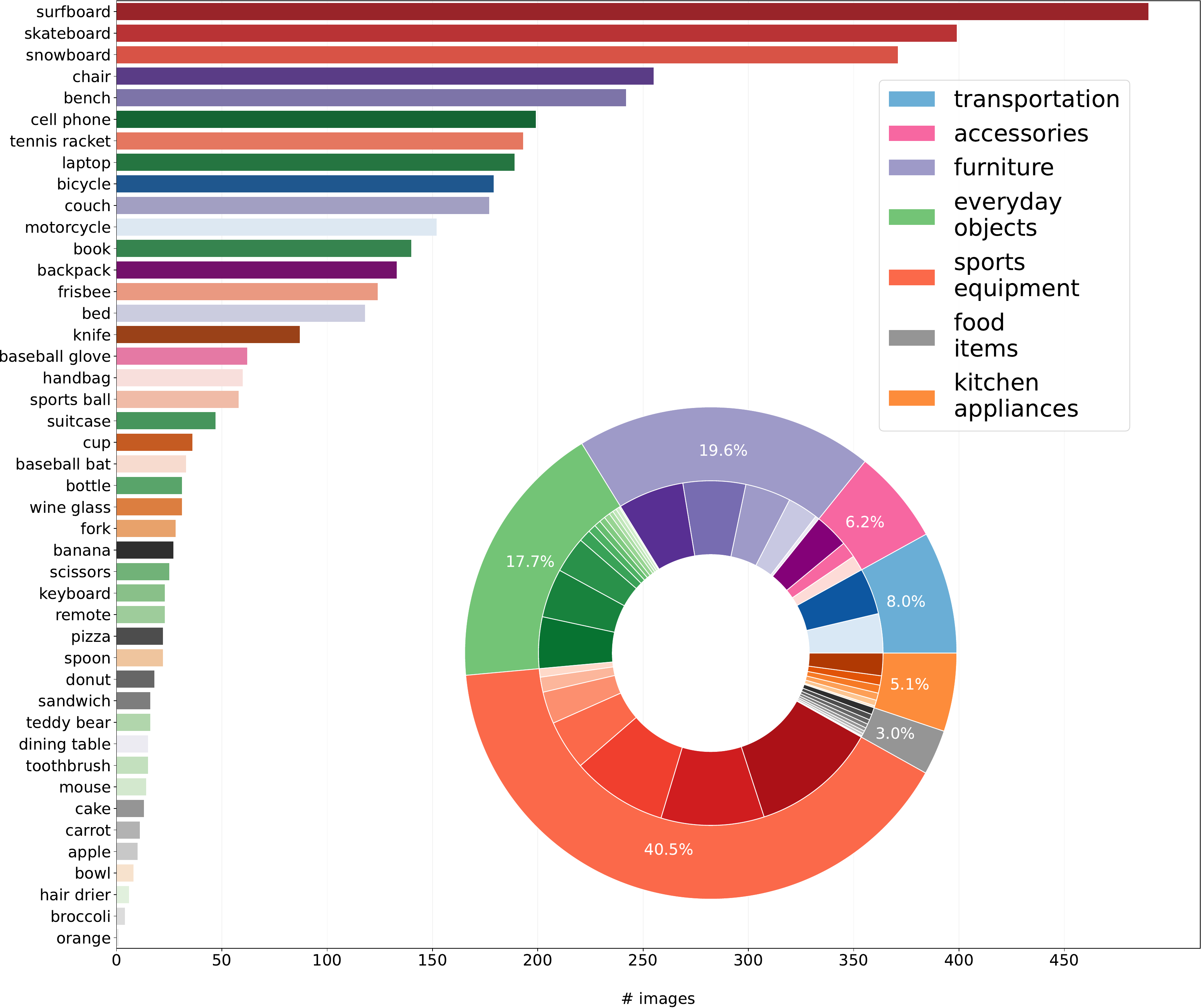}
    \caption{
            Statistics on the %
            object categories in \nameDataset.
            \textbf{Histogram:} 
            Object labels (y-axis) and the number of images in which they are present (x-axis).
            \textbf{Pie chart:} 
            Object labels are grouped into 7 main categories; inner colors correspond to the colors in the histogram.
    }
    \label{fig:supmat:object_categories}
\end{figure*}

\nameDataset contains 4123 images with paired human and object 3D contact. The images span 44 object categories.
This is fewer than the 69 object categories in \DAMON as we identify and reject object categories that are never (or rarely) in contact with humans in images, such as a wall clock, fire hydrant, plant, TV, etc. Additionally, we exclude objects that are too large, so they are severely 
truncated in images %
such as (sitting in an) airplane, boat, car, bus, train, etc.
We also filter out images of children, since their smaller size would ``compromise'' contacts annotated on the bigger default-shape \smpl body. 
For the complete list of included object categories and their distribution refer to \cref{fig:supmat:object_categories}. \camready{Note that we use the same train, validation and test splits as the \DAMON dataset.}

\begin{figure}
    \centering
    \includegraphics[width=1.0 \linewidth]{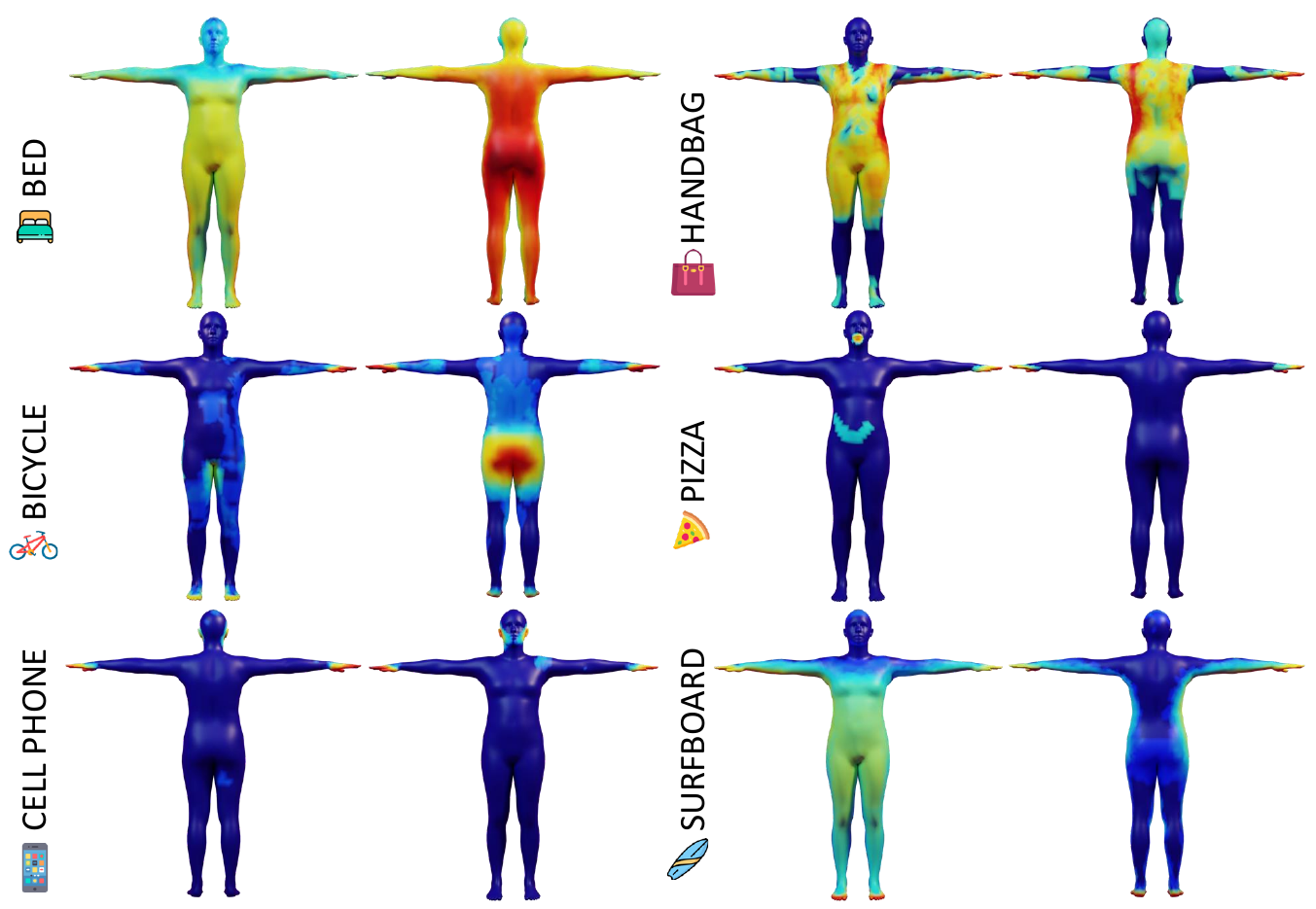}
    \caption{
            Aggregate statistics showing object-wise contact probabilities across all body vertices in the \nameDataset dataset. The body part closeups show the contact probabilities normalized for that body part. \textcolor{red}{Red} implies higher probability of contact while \textcolor{blue}{blue} implies lower probability. \faSearch~\textbf{Zoom in}.
    }
    \label{fig:supmat:objwise_probability}
\end{figure}

\camready{In Fig.~\cref{fig:supmat:objwise_probability}, we present the aggregate vertex-level contact distributions for six object categories: bed, bicycle, cell phone, handbag, pizza, and surfboard. These distributions illustrate that our dataset captures a wide range of interaction patterns, reflecting both frequent (canonical) and infrequent (rare or edge-case) usage scenarios. This diversity highlights the richness of human-object interactions in our dataset}

\subsection{Quality control}

We adopt several strategies to ensure high-quality annotations in \nameDataset. First, we select high-performing AMT annotators through a rigorous two-part qualification process. 
Specifically, annotators are required to
(i)  watch a detailed tutorial video (see %
\video), and 
(ii) complete test annotations on a standardized set of 10 sample images. 
We evaluate the annotator responses by computing the point-to-point Euclidean distance between their annotations and author-annotated \emph{pseudo-ground-truth} (pseudo GT) labels, as performed in \DAMON~\cite{tripathi2023deco}. With this, we qualify 17 out of 150 participants. 
Second, we release annotation tasks in small batches and visually inspect the quality of contact annotations per batch. Annotations flagged as incorrect or low-quality are repeated in the next batch.

\section{Leveraging VLLMs in \nameMethod}
\label{supmat:vllm_contact_estimation}

In this work, we exploit the general world knowledge of VLLMs in two ways (i) to initialize object scale in \nameMethod and (ii) to refine human-contact predictions from \DECO. 

For initializing object scale, we input the test image to GPT-4V and query the object's scale by using the following prompt:
\begin{tcolorbox}[colback=gray!10, colframe=gray!50, sharp corners, boxrule=0.5mm]
\footnotesize
\texttt{How big is the <OBJECT> in the <IMAGE> that the human is interacting with? Use the other objects and the scale of the human to estimate the size. Answer should be single number, in meters, that corresponds to the length of the longest side of the <OBJECT>.}
\end{tcolorbox}

We also use \mbox{GPT-4V} to refine \DECO's body contact predictions. 
\camready{Since \nameMethod relies on \DECO predictions 
to retrieve contacts from \nameDataset
for both the body and the object, as well as the object shape, any errors in the estimated body contact may propagate to subsequent steps in the \nameMethod %
pipeline.} 

While \DECO is robust and generalizes well to in-the-wild scenarios, it has a strong bias for predicting false positives on the feet, and often misses body parts in contact. To tackle this, we refine \DECO's predictions by removing feet contact 
if it is not predicted by GPT-4V, and adding contact on any body parts that are additionally predicted by \mbox{GPT-4V}. To this end, we use the following prompt:

\begin{tcolorbox}[colback=gray!10, colframe=gray!50, sharp corners, boxrule=0.5mm]
\footnotesize
\texttt{List the body parts of the human that are in contact with the <OBJECT> (touching or supporting the object) in this <IMAGE>. \\
These are all the body parts to consider: head, neck, torso, hips, leftUpperArm, rightUpperArm, leftForeArm, rightForeArm, leftHand, rightHand, leftUpperLeg, rightUpperLeg, leftLowerLeg, rightLowerLeg, leftFootSole, rightFootSole, topOfLeftFoot, topOfRightFoot. Answer should be only a comma-separated list of the body parts, nothing else.}
\end{tcolorbox}

\begin{figure}
    \centering
    \includegraphics[width=1.0 \linewidth]{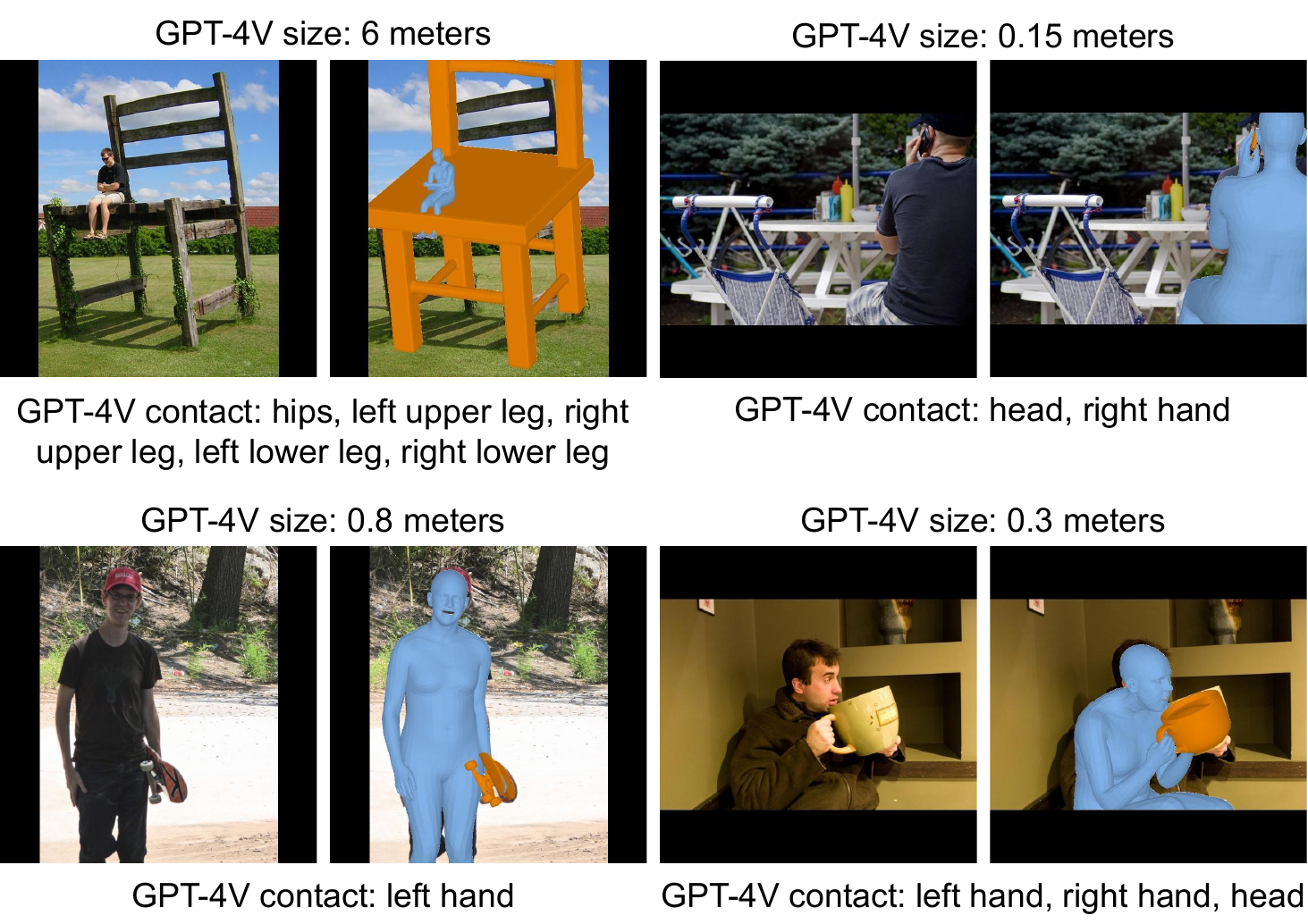}
    \caption{\camready{Visual examples showing \GPT predicted object size and contacting body parts. ``Size'' implies length of the largest dimension.}}
    \label{fig:supmat:gpt_impact}
\end{figure}

\zheading{Impact of \GPT}
\camready{DECO+\GPT contact is higher quality relative to \DECO; 
F1 improves from 0.29 to 0.35 on \InterCap. PA-CD$_{\human+\object}$ also improves, from 11.76 to 10.33. On object scale, \GPT yields 17.0 cm RMSE on \InterCap}. \cref{fig:supmat:gpt_impact} shows visual examples indicating the estimated size and contacting body parts from \GPT on some challenging images with diverse objects. The object size refers to the length of the largest dimension.

\section{\nameMethod additional details}
\label{supmat:evaluation_details}

\subsection{Implementation details}

We use the Adam optimizer, with parameter-specific learning rates ranging between 0.01 and 0.04, assigning higher rates to object rotation parameters to accelerate exploration of their %
search space. 

Empirically, we find that reconstructions are most consistent with the following set of loss weights: 
\begin{itemize}
\item 
for the second stage of the optimization we use weights: 
$\lambda_\contact = 4, \quad
\lambda_\collision = 100, \quad
\lambda_\object^\mask = 0.4, \quad
\lambda_\object^\scale = 4 $
\item
for the third stage of the optimization we use weights: 
$\lambda_\contact = 4, \quad
\lambda_\collision = 50, \quad
\lambda_\human^\mask = 0.1, \quad
\lambda_{\theta_\mathcal{C}} = 0.05$
\end{itemize}

\subsection{Quantitative evaluation details on \InterCap}

We evaluate on \InterCap by reporting the Procrustes-Aligned (PA) Chamfer Distance (CD). Since state-of-the-art methods use different output formats, we standardize to ensure a fair evaluation.

\camready{
While using the joint human and object mesh for alignment is standard practice~\cite{nam2024contho}, the Procrustes-alignment algorithm assumes higher weight on the human, since the human mesh has considerably more vertices than the object mesh in 3D \hoi datasets~\cite{huang2022intercap, bhatnagar2022behave}. As a side-effect, this leads the PA-CD$_\human$ to be often lower than PA-CD$_\object$, which is evident in Tab. \textcolor{red}1 in main.}

For evaluating \CONTHO~\cite{nam2024contho} we use the authors' published code and annotation file. \camready{We adapt their evaluation code for all methods and use the same \InterCap test split they release.}

\PHOSA~\cite{zhang2020phosa} outputs SMPL meshes to represent the human pose and shape, which are inconsistent with the ground-truth SMPL-X meshes in \InterCap. While we use the joint human and object mesh to Procrustes align predictions with ground-truth, in case of the human body, we exclude the head vertices. We do this as the body vertices share the same topology between SMPL and SMPL-X, whereas the head does not.
After alignment, we sample the same number of points in both \PHOSA predicted meshes as well as the ground-truth meshes in \InterCap to compute chamfer distance. 

\camready{
The
\HDM~\cite{xie2023tprocigen}
model trains on
\Procigen~\cite{xie2023tprocigen}, a synthetic dataset 
building on \BEHAVE and \InterCap.}
It outputs point clouds for both the human and the object in \InterCap's ``Cam1'' coordinate frame. To compute the chamfer distance, we bring all ground-truth meshes from \InterCap in the same \camready{``Cam1''} coordinate frame and sample equal number of points ($=8192$) as in the predictions, \camready{before aligning the predicted and ground-truth point clouds using Iterative Closest Point (ICP) and computing CD. Note that for alignment, we use the \emph{combined} human and object point cloud. Unlike for Procrustes-alignment, since ICP uses the same number of points from the human and the object, \HDM's chamfer distance scores are more balanced between the human and the object in Tab. \textcolor{red}1 in main.}

\subsection{Perceptual study details}

\begin{figure}
    \centering
    \vspace{-0.5 em}
    \includegraphics[width=0.93 \linewidth]{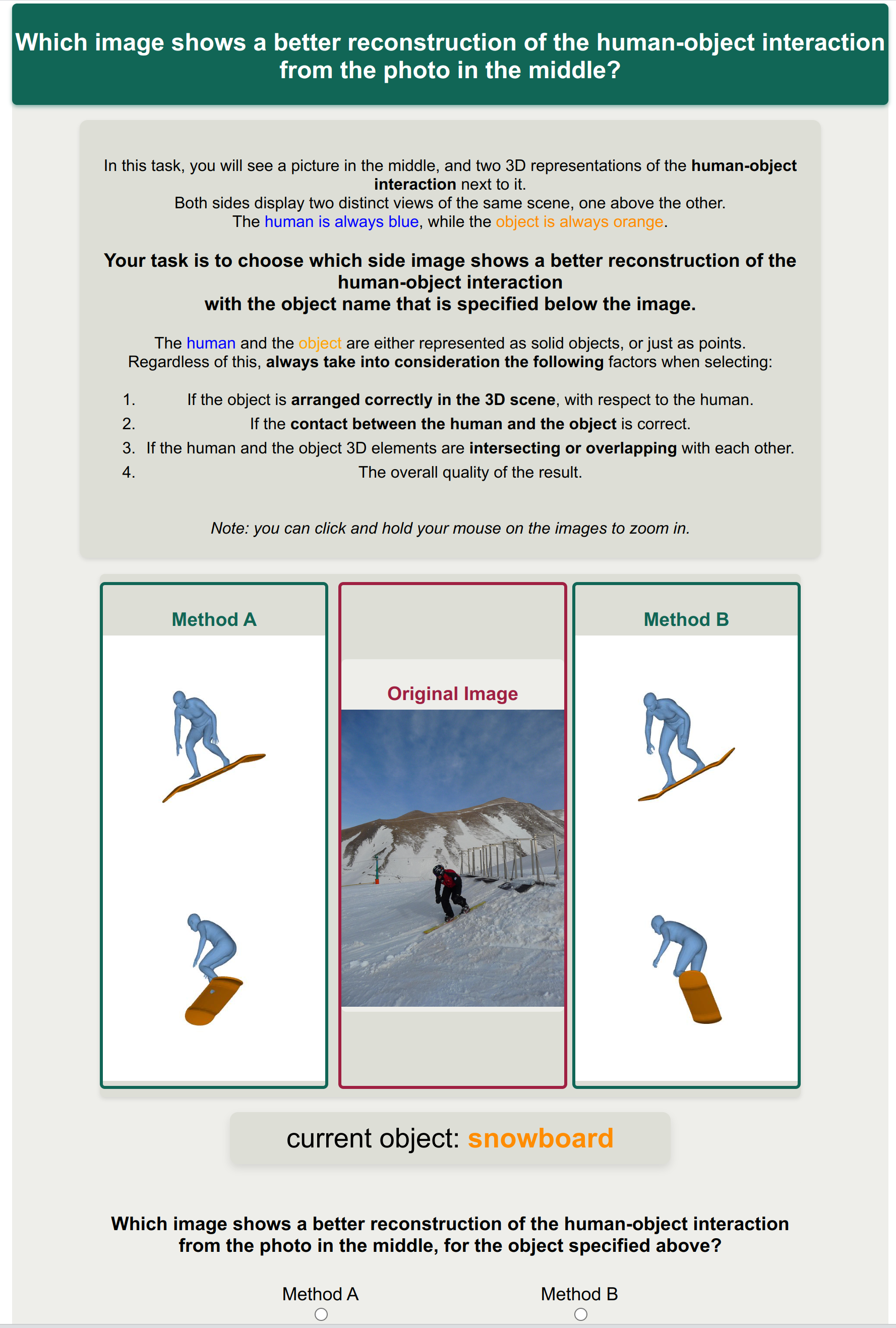}
    \vspace{-0.5 em}
    \caption{
                        Layout of the perceptual study. Below the extensive but simple instructions, participants are presented with two different views of the reconstructions from two methods (randomly swapped). 
                        Our interface correctly adapts to any screen size, but users are also able to click on the images to zoom in
    }
    \label{fig:supmat:perceptual_study}
    \vspace{-0.5 em}
\end{figure}

To ensure reliable participants, we repeat the first three images at the end of the study and exclude them during evaluation to serve as a warm-up for participants.
We also include four catch trials—pairs of images with decisions that are intentionally straightforward—to identify and filter out participants who may provide random inputs. We exclude all submissions with even a single failed catch trial, which results in discarding 27 out of 100 total completions.
For the layout we use in the perceptual study, see \cref{fig:supmat:perceptual_study}.

\subsection{Additional Ablations}

\begin{table}
\centering
\resizebox{1.00 \columnwidth}{!}{
\begin{tabular}{l|cccc|ccc}
\Xhline{3\arrayrulewidth}
\multicolumn{1}{c|}{\bf Stage IDs} & $\boldsymbol{\mathcal{L}_\contact}$ & $\boldsymbol{\mathcal{L}_{\object,\mask}}$ & $\boldsymbol{\mathcal{L}_\collision}$ & $\boldsymbol{\mathcal{L}_{\human,\mask}}$ & \bf CD$_\human$ $\downarrow$ & \bf CD$_\object$ $\downarrow$ & \bf CD$_{\human+\object}$ $\downarrow$ ~\\ 
\cline{1-8}
1 + 2 + 3     & \xmark & \cmark & \cmark & \cmark & 22.71 & 39.39 & 26.63 \\ \hline
2 + 3 (no 1)  & \cmark & \cmark & \cmark & \cmark & 9.24 & 34.39 & 12.9  \\ \hline
1 + (2\&3 comb.) & \cmark & \cmark & \cmark & \cmark & 8.13 & 19.1 & 10.66  \\ \hline
1 + 2 + 3 (\nameMethod)     & \cmark & \cmark & \cmark & \cmark & 6.66 & 13.34 & 8.36 \\ \hline
\end{tabular}
}
\vspace{-1 em}
\caption{
            Additional ablations, extending \mbox{Tab.~\textcolor{red}{2}} of the paper. 
}
\label{table:rebuttal:extra_ablation}
\vspace{-1.2 em}
\end{table}

\camready{We ablate \refine{$\boldsymbol{\mathcal{L}_\contact}$ in \cref{table:rebuttal:extra_ablation}}, top and bottom rows. Note that \nameMethod needs both human and object contact maps 
for $\boldsymbol{\mathcal{L}_\contact}$, and hence, we cannot ablate them separately. Results show that $\boldsymbol{\mathcal{L}_\contact}$ is essential for performance.}

\camready{
Next, we analyze alternative optimization strategies for \nameMethod. To evaluate the effect of Stage 1 which uses dense contact correspondences to initialize object pose \wrt the body, we run only Stage 2 and 3 and report results in \cref{table:rebuttal:extra_ablation}, second row. In \cref{table:rebuttal:extra_ablation}, third row, we first run Stage 1, followed by a joint optimization of Stage 2 and 3 together. The results show that the proposed optimization scheme in \nameMethod significantly outperforms these alternatives, particularly for recovering accurate object poses.}

\begin{figure}
    \centering    %
    \includegraphics[trim=000mm 000mm 000mm 000mm, clip=true, width=1.0 \linewidth]{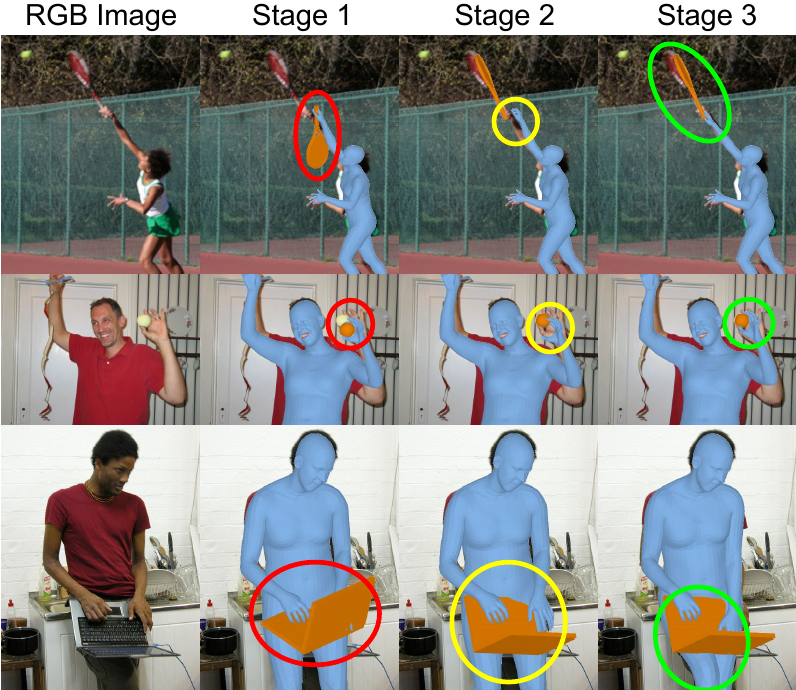} 
    \vspace{-2.0 em}
    \caption{
                Ablation study for \nameMethod's 
                stages. 
    }
    \label{fig:experiments:pico_ablation}
\end{figure}

\camready{
The qualitative ablation in \cref{fig:experiments:pico_ablation} demonstrates the effect of each stage in \nameMethod. In Stage 1, \nameMethod establishes contact between the human and the object, though the object may not yet be aligned with the image. Stage 2 refines the object's alignment with the image, albeit at the cost of slight contact misalignment. Finally, Stage 3 optimally balances contact, interpenetration, and image alignment by refining the human’s contacting limbs. This results in a 3D \hoi that is both image-aligned in 2D and plausible in 3D.
}

\subsection{Failure cases}

\begin{figure}
    \centering
    \includegraphics[width=0.99 \linewidth]{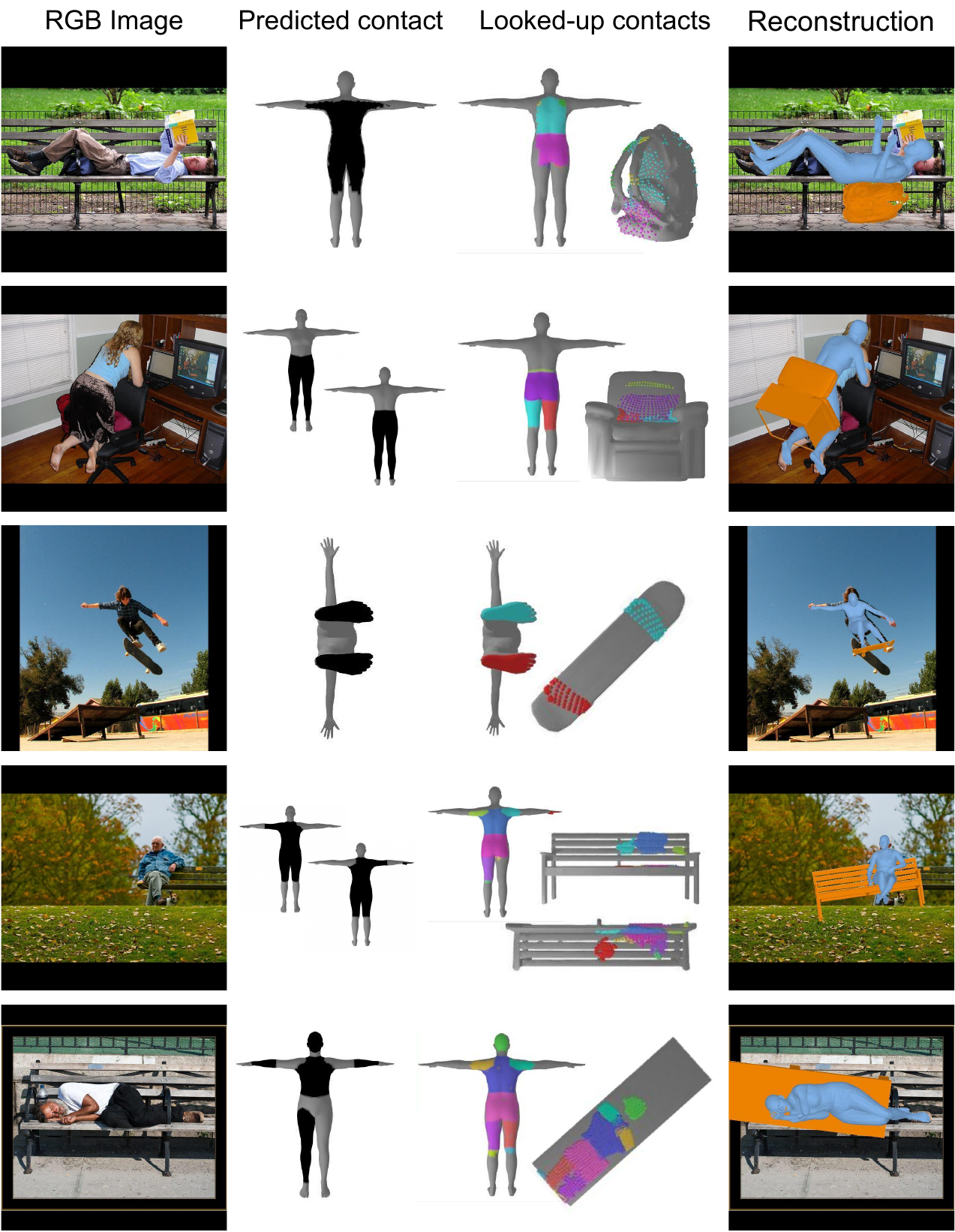}
    \caption{
            Example interactions where \nameMethod lookups on \nameDataset fail. Each row from left to right: 
        input image, predicted body contact from \DECO + GPT-4V, looked-up contact from \nameDataset and 3D reconstructions overlaid on the images. Rows 1-3: incorrect human contact prediction. Rows 4-5: incorrect object contact retrieval.
    }
    \label{fig:supmat:failcases_lookup}
\end{figure}

\begin{figure}
    \centering
    \includegraphics[width=0.99 \linewidth]{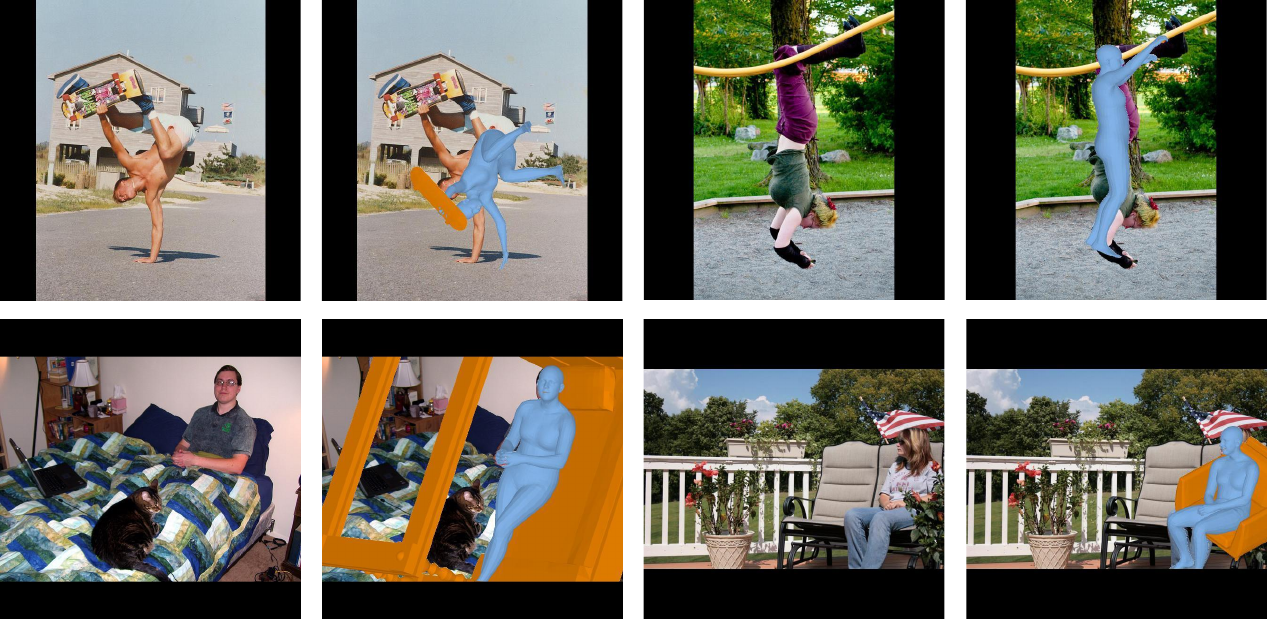}
    \caption{
            Failure cases of \nameMethod. Each row (from left to right) shows two input images and corresponding \nameMethod reconstructions overlaid on the image. 
            Top row: incorrect human pose initialization. Bottom row: incorrect object retrieval.
    }
    \label{fig:supmat:failcases_others}
\end{figure}

\camready{Like all current methods, \nameMethod might fail under truly novel interactions if these differ significantly from those included in \nameDataset. We show examples of \nameMethod failures under unusual contact scenarios in \cref{fig:supmat:failcases_lookup}, both due to inaccurate human contact (row 1-3) and object contact (row 4-5). \Cref{fig:supmat:failcases_others} demonstrates additional \nameMethod failures caused due to incorrect human pose initialization (row 1) and incorrect object retrieval (row 2). Further, to develop understanding of \nameMethod failures, we randomly sampled 500 \nameMethod reconstructions and hired Master's students to categorize them into failure modes. Most \nameMethod failures result from:
(1) Incorrect human pose initialization by \OSX~\cite{lin2023osx} ($5/500$),
(2) incorrect object retrieval which does not match the image ($12/500$), 
(3) incorrect human-contact prediction by \DECO ($85/500$), and 
(4) invalid object contact retrieval even when the inferred human contact is correct ($20/500$). }

\camready{
Despite the failures, we note that \nameMethod handles significantly more object instances than existing work; 
\nameMethod handles 627 objects, 
namely 
1-2 orders of magnitude more 
than the 8 and 30 objects of \PHOSA and \CONTHO/\HDM, respectively. Further, as shown in 
\mbox{Tab.~\textcolor{red}{1}} and 
\mbox{Figs.~\textcolor{red}{7}} and 
\textcolor{red}{8} in main, 
\nameMethod achieves SOTA performance on OOD and in-the-wild datasets, indicating superior generalization.} 

\section{Additional qualitative results}
\label{supmat:pico_fit_extra}

\begin{figure*}
    \centering
    \vspace{-0.5 em}
    \includegraphics[width=1.0 \linewidth]{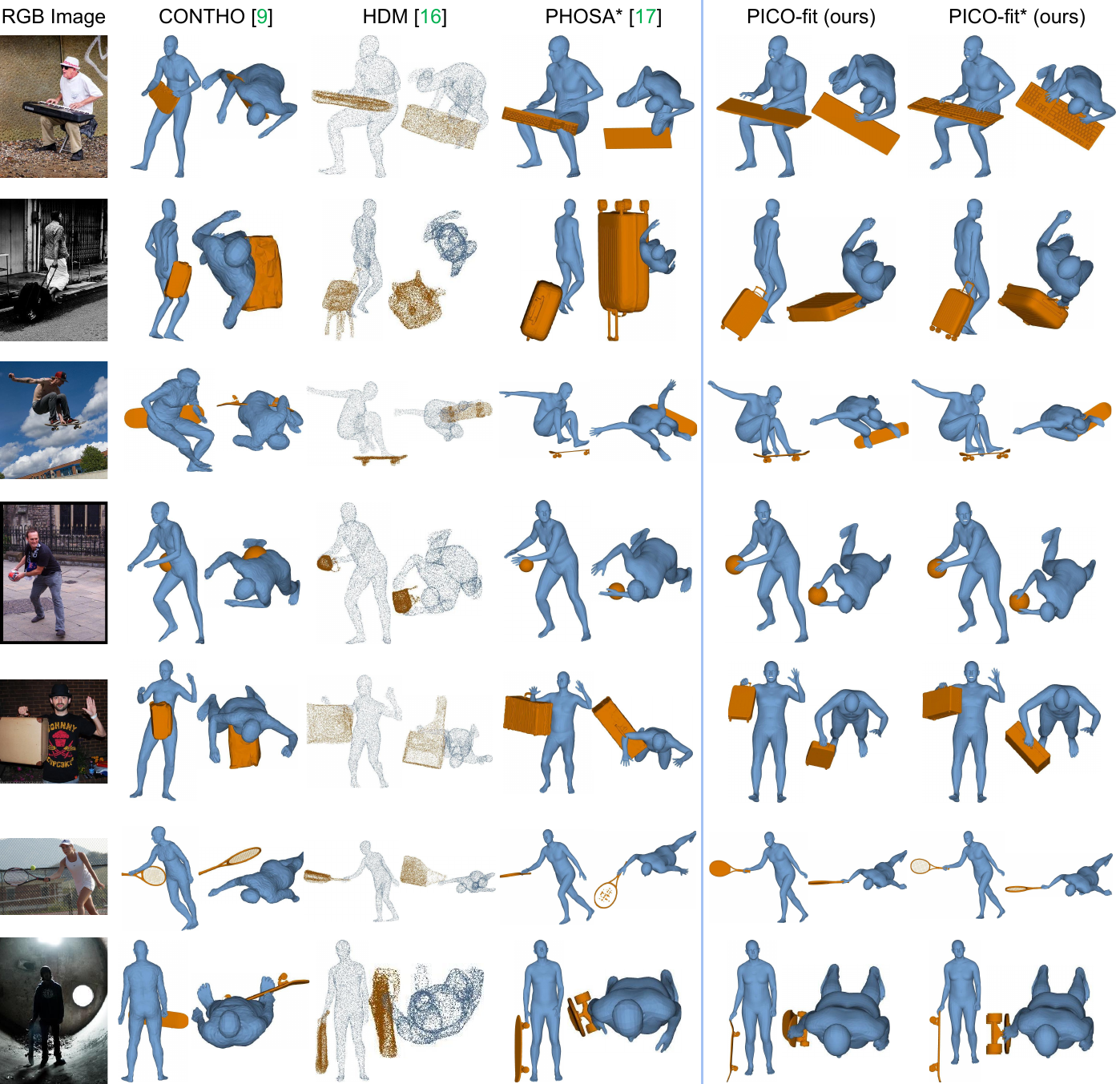}
    \vspace{-2.0 em}
    \caption{
                        Qualitative evaluation of \CONTHO, \HDM and $\text{\PHOSA}^*$ alongside \nameMethod and \nameMethodStar on object categories handled by all baselines. Since \HDM cannot produce image overlays, we present only front- and top-down views for all methods
    }
    \label{fig:supmat:sota_compare}
\end{figure*}

\begin{figure*}
    \centering
    \includegraphics[width=1.0 \linewidth]{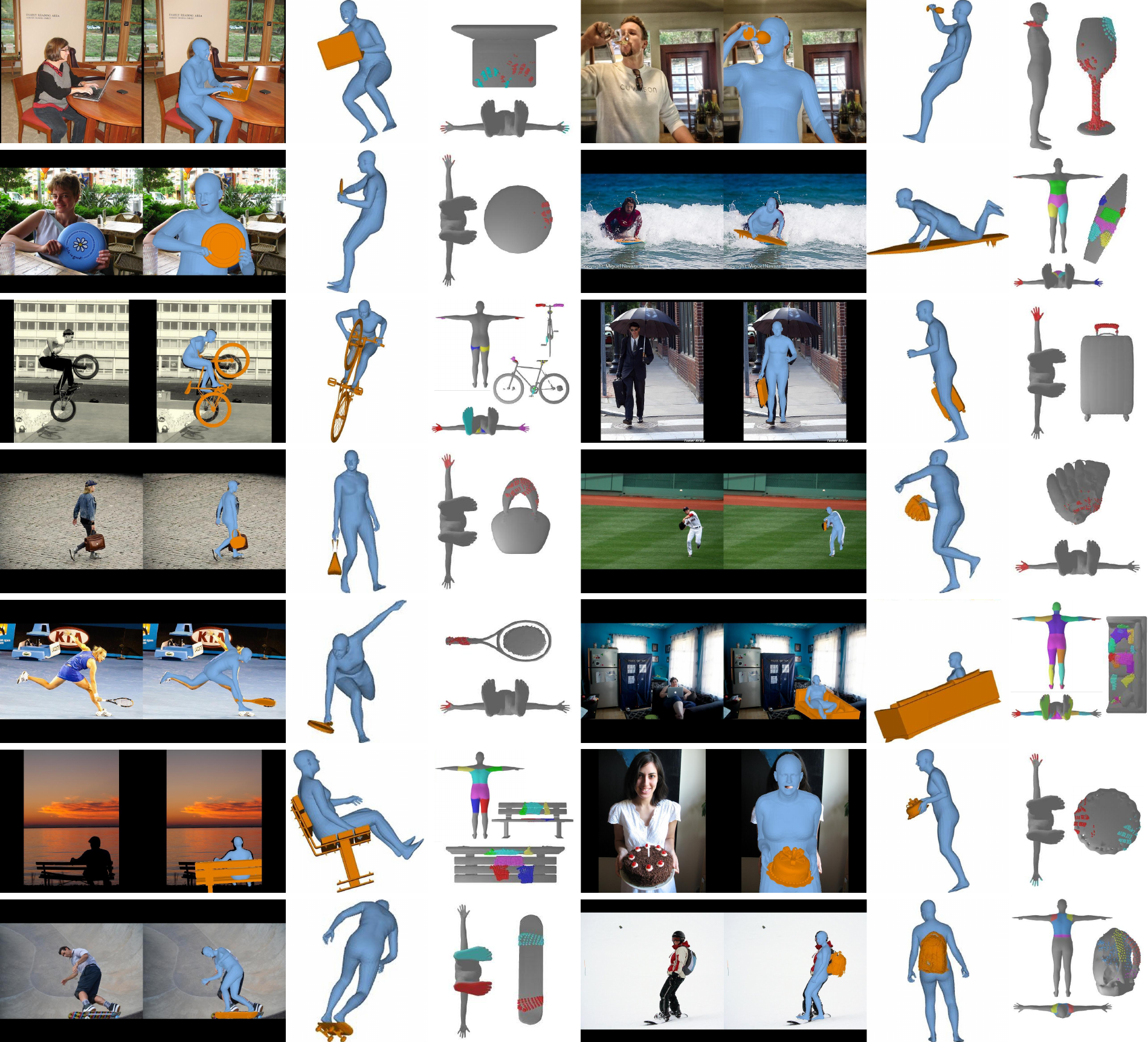}
    \caption{
                        3D \hoi reconstructions from our \nameMethod method on various object categories. 
                        For each triplet in each row we see (from left to right): an input RGB image, \nameMethod's estimated meshes overlaid on the image (camera view), a side view, \camready{and the contact annotations that were looked up and taken from \nameDataset.}
    }
    \label{fig:supmat:pico_lookup_examples}
\end{figure*}

\begin{figure*}
    \centering
    \includegraphics[width=1.0 \linewidth]{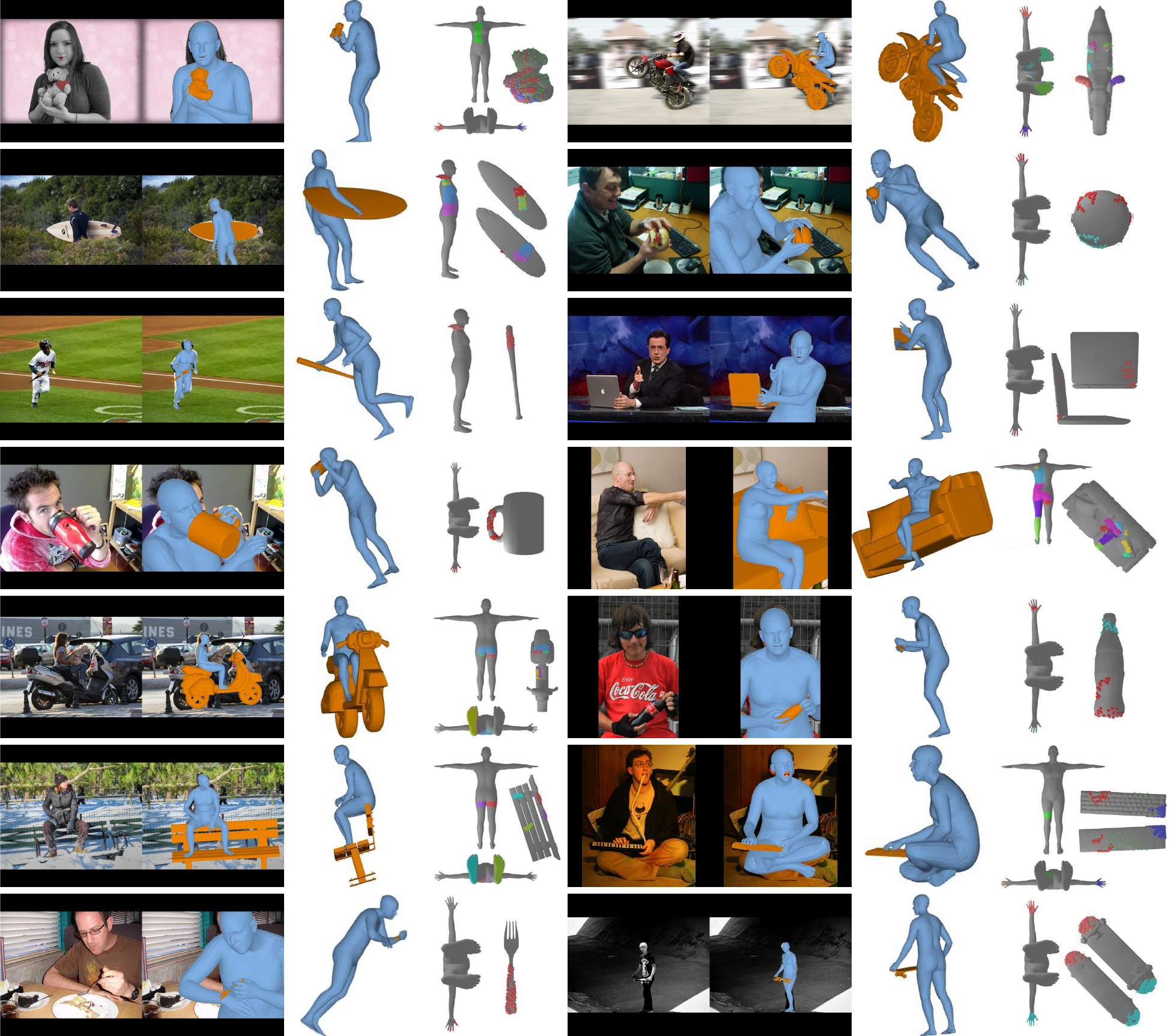}
    \caption{
                        3D \hoi reconstructions from our \nameMethodStar method on various object categories. 
                        For each triplet in each row we see (from left to right): an input RGB image, \nameMethodStar's estimated meshes overlaid on the image (camera view), a side view, \camready{and the corresponding contact annotations from \nameDataset.}
    }
    \label{fig:supmat:pico_star_examples}
\end{figure*}

\cref{fig:supmat:sota_compare} shows qualitative comparisons of \CONTHO, \HDM and $\text{\PHOSA}^*$ alongside \nameMethod and \nameMethodStar for object categories handled by all baselines.
\cref{fig:supmat:pico_lookup_examples} shows \hoi reconstructions from \nameMethod on various object categories, and \cref{fig:supmat:pico_star_examples} does the same for \nameMethodStar.

\end{document}